\documentclass[letterpaper]{article} 
\usepackage{aaai2026}  
\usepackage{times}  
\usepackage{helvet}  
\usepackage{courier}  
\usepackage[hyphens]{url}  
\usepackage{graphicx} 
\usepackage{natbib}  
\usepackage{caption} 
\usepackage{graphicx}
\usepackage{amsmath}
\usepackage{algorithm}
\usepackage{algorithmic}
\usepackage{booktabs} 
\usepackage{multirow}
\usepackage{amssymb}
\usepackage{adjustbox}
\usepackage{listings}
\usepackage{subcaption}
\usepackage{enumitem}
\usepackage[most]{tcolorbox}
\usepackage{newfloat}
\usepackage{listings}
\DeclareCaptionStyle{ruled}{labelfont=normalfont,labelsep=colon,strut=off} 
\floatstyle{ruled}
\newfloat{listing}{tb}{lst}{}
\floatname{listing}{Listing}
\newcommand{\sdataset}{SLQA}
\newcommand{\ldataset}{SEQA}
\newcommand{\wdataset}{WTQ}
\newcommand{\hdataset}{HybQA}
\newcommand{\tdataset}{TabFact}
\newcommand{\model}{\textsc{TaDRe}}
\newcommand{\smodel}{Decomposition Planner}
\newcommand{\dmodel}{Table Decomposer}
\newcommand{\amodel}{Answer Generator}
\newcommand{\vmodel}{Decomposition Refiner}

\title{Towards Question Answering over Large Semi-structured Tables}
\author{
    Yuxiang Wang,
    Junhao Gan,
    Jianzhong Qi
}
\affiliations{
    School of Computing and Information Systems, The University of Melbourne\\

    yuxiang.wang8@student.unimelb.edu.au, \{ junhao.gan, jianzhong.qi\}@unimelb.edu.au
}

\usepackage{bibentry}

\begin{document}

\maketitle

\begin{abstract}
Table Question Answering (TableQA) attracts strong interests due to the prevalence of web information presented in the form of semi-structured tables. Despite many efforts, TableQA over large tables remains an open challenge. This is because large tables may overwhelm models that try to comprehend them in full to locate question answers.   
Recent studies reduce input table size by decomposing tables into smaller, question-relevant sub-tables via generating programs to parse the tables.  However, such solutions are subject to program generation and execution errors and are difficult to ensure decomposition quality. To address this issue, we propose \model, a TableQA model that incorporates both pre- and post-table decomposition refinements to ensure  table decomposition quality, hence achieving highly accurate TableQA results. 
To evaluate \model, we construct two new large-table TableQA benchmarks via LLM-driven table expansion and QA pair generation. Extensive experiments on both the new and public benchmarks show that \model\ achieves state-of-the-art performance on large-table TableQA tasks.
\end{abstract}


\section{Introduction}
Table Question Answering (TableQA) aims to answer natural language questions grounded in semi-structured tabular data. It is an important problem because not only is information on the web often presented in the form of semi-structured tables (just ``tables'' hereafter), but also solving the problem helps to understand the capability of language models---which are usually trained with unstructured texts---to comprehend semi-structured data.

\begin{figure}[t]
    \centering
    \begin{subfigure}[c]{0.43\textwidth}
        \centering
        \includegraphics[width=1\linewidth]{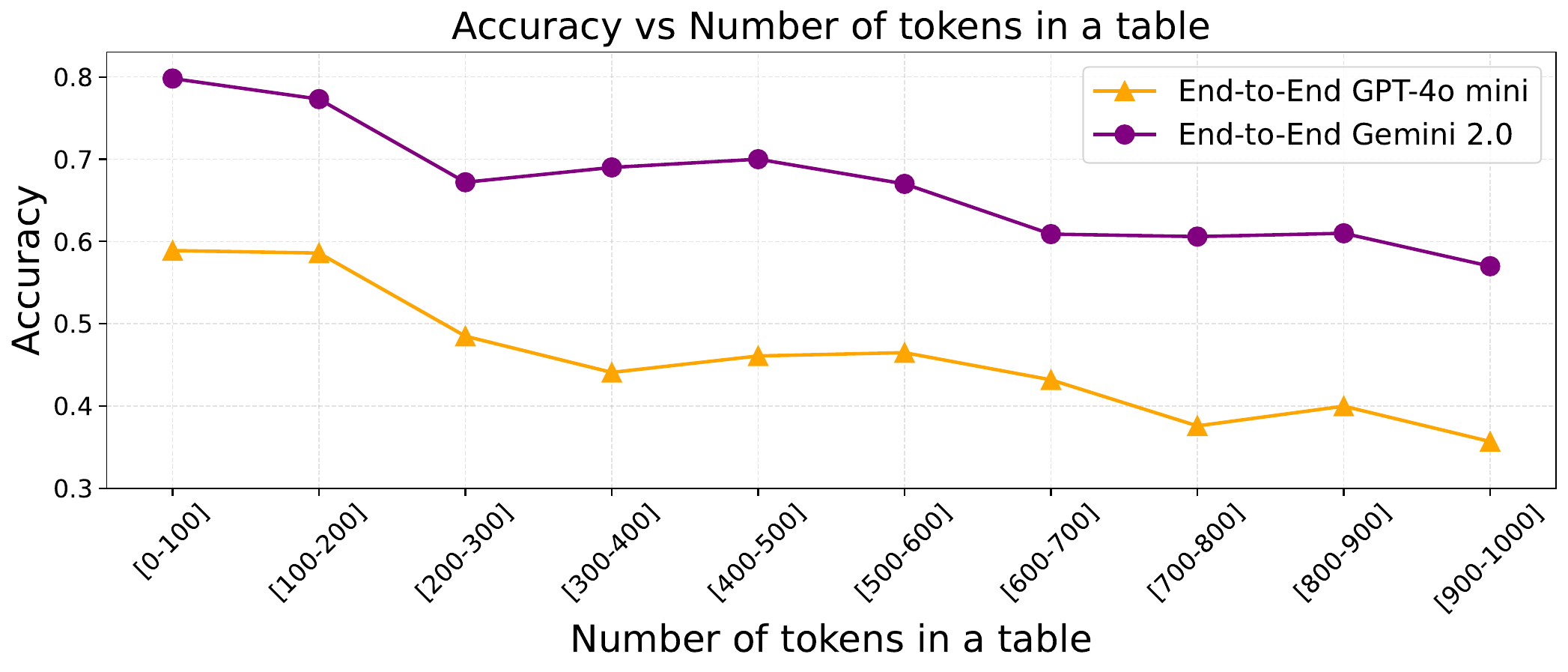}
        \caption{}
        \label{fig:token_acc}
    \end{subfigure}
    \begin{subfigure}[c]{0.45\textwidth}
        \centering
        \includegraphics[width=1\linewidth]{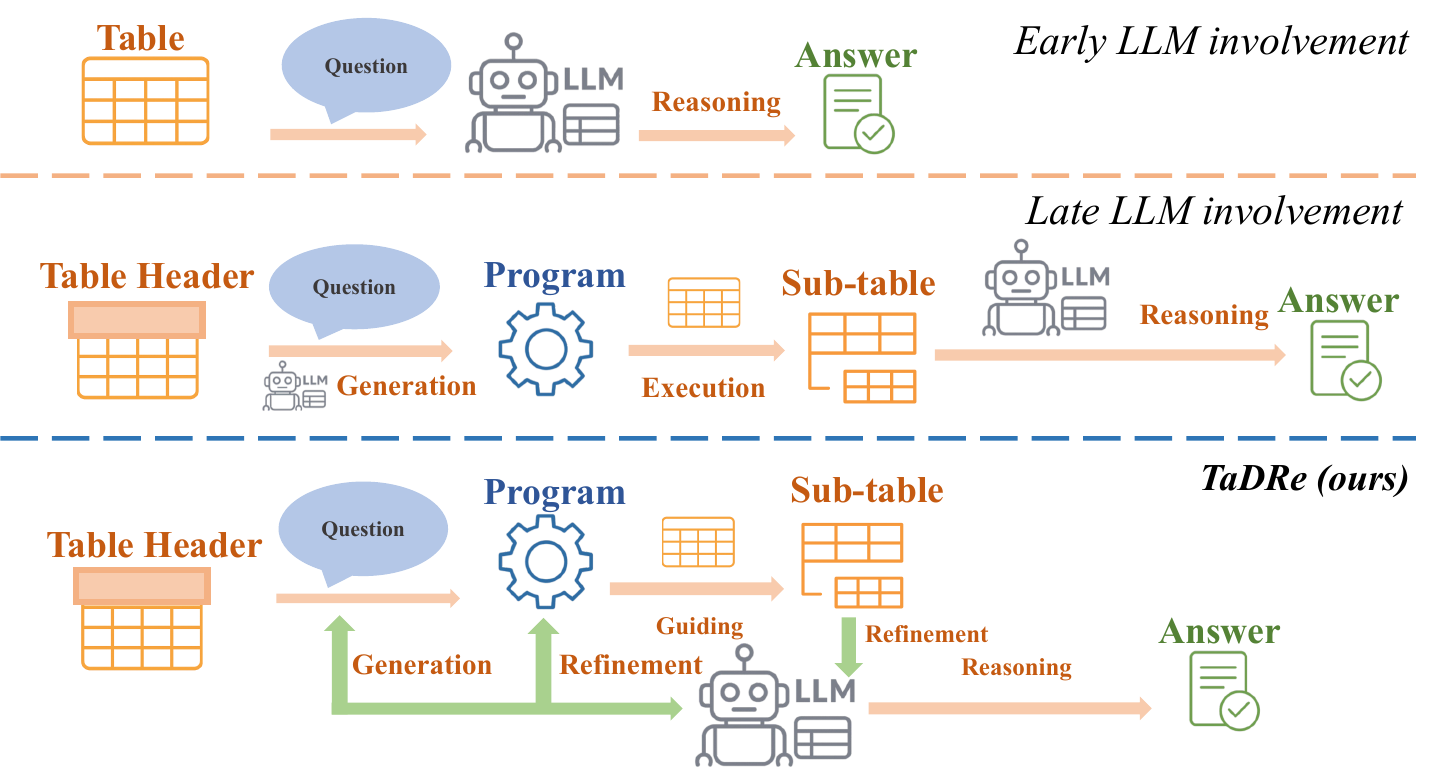}
        \caption{}
        \label{fig:llm_involve}
    \end{subfigure}
    \caption{(a) TableQA accuracy of running \texttt{GPT4-o mini} and \texttt{Gemini 2.0} on tables of different lengths (i.e., numbers of tokens) in the  \texttt{WikiTableQuestions} dataset~\cite{pasupat-liang-2015-compositional}. (b) Illustration of different LLM involvement strategies in TableQA.}
    \label{fig:combined}
\end{figure}

Recent studies~\cite{cheng2023binding, liu-etal-2024-rethinking, wang2025accurate} exploit Large Language Models (LLMs) to comprehend both natural language questions and tables and generate answers. Such solutions are challenged by large tables~\cite{LiuLHPBPL24}, e.g., tables with just over 500 tokens (cf.~Figure~\ref{fig:token_acc}), which could contain too much irrelevant information that overwhelms LLMs.


Studies on TableQA over large tables generally take a decompose-then-reason approach that first decomposes a large table into smaller, question-relevant sub-tables, upon which a model (e.g., an LLM) reasons to generate the question answer. Some existing works~\cite{ye2023large, lee2024learning, wang2024chainoftable} feed an entire table to an LLM for decomposition, where the LLM is again challenged by the large table size. Others ~\cite{pourreza2023, ChenME00CFLLP24, NahidR24} use LLMs to analyze the input question and table headers, and generate table parser programs which are executed for table decomposition. Due to the semi-structured nature of the tables and the generative nature of the table parser programs, the programs may fail to fetch question-relevant sub-tables or may not run at all. 

We observe that these existing solutions run LLMs on tables either \emph{too eagerly} (i.e., read full tables for decomposition) or \emph{too late} (i.e., wait until tables are decomposed by a parser program), which leaves a gap to fill -- see different pipelines in Figure~\ref{fig:llm_involve}, and an opportunity to enhance TableQA accuracy over large tables.



We propose a TableQA model named \model\ where LLMs are an active participant in the table decomposition step, i.e., \underline{ta}ble \underline{d}ecomposition with LLM \underline{re}soning.

\model~first employs a \emph{\smodel}, an LLM-based component that leverages few-shot in-context learning to generate SQL queries. The queries are structured reasoning paths for question-relevant sub-table retrieval. Rather than executing these queries directly, a table semantics- and structure-aware \emph{\dmodel} parses them to extract semantic signals, such as column names, values, and conditions, which are then converted into a Python program via a SQL-to-Python semantic mapping to extract a question-relevant sub-table.
This yields a compact and relevant sub-table for answering the input question. 

To enhance decomposition robustness, we introduce an LLM-based \emph{\vmodel}, which operates both before and after table decomposition. Before decomposition, it verifies and corrects the generated SQL queries to ensure alignment with the input table structure and question semantics. After decomposition, we apply a novel, Chain-of-Table-Retrieval (CoTR) mechanism, which dynamically augments the sub-tables when they 
lack information to answer the input question. This design provides finer-grained control, avoids unnecessary token overhead, and adapts better to large tables.  Finally, the refined sub-table is passed to an \emph{\amodel} (another LLM) to produce the answer.

To support evaluation for large-table TableQA, we construct two new datasets derived from the \texttt{Spider} dataset~\cite{yu-etal-2018-spider}: (1) Spider Large TableQA Dataset (denoted as \texttt{\sdataset}), built by generating natural language QA pairs over original large tables, and (2) Spider Expanded TableQA Dataset (denoted as \texttt{\ldataset}), which expands small tables from \texttt{Spider} into large ones using LLM-based data augmentation, followed by QA pair generation. We also extract large tables from the \texttt{WikiTableQuestions} dataset \cite{pasupat-liang-2015-compositional} and the \texttt{HybridQA} dataset \cite{ChenZCXWW20} to enrich our evaluation. 


This paper makes the following contributions:
\begin{itemize}[leftmargin=*, noitemsep, topsep=0pt]
\item We propose \model, a novel TableQA model designed for large semi-structured tables. \model~transforms SQL queries into a decomposition reasoning step and integrates a structure- and semantics-aware \dmodel~to extract sub-tables without executing the SQL queries.

\item We introduce \vmodel, a refinement mechanism that operates both before and after table decomposition. It verifies SQL quality and supports dynamic sub-table augmentation via a Chain-of-Table-Retrieval (CoTR) mechanism.

\item We present \texttt{\ldataset} and \texttt{\sdataset}, two large-table TableQA benchmarks automatically constructed from the \texttt{Spider} dataset via LLM-driven table expansion and QA generation. These datasets contain natural language questions over large tables and will be released with this paper.

\item Extensive experiments show that \model~achieves state-of-the-art (SOTA) performance on large-table TableQA tasks while generalizing well to small-table benchmarks, highlighting its efficiency and robustness.
\end{itemize}

\section{Related Work}
Existing TableQA studies can be broadly categorized into two groups based on whether they explicitly separate table decomposition from  semantic reasoning:
(1)~\textit{single-stage models} and (2) \textit{decompose-then-reason models}.


\paragraph{Single-Stage Models} Single-stage models jointly encode the full input table and question to generate answers without constructing intermediate sub-tables. 

\underline{PLM fine-tuning.} Early studies fine-tune Pre-trained Language Models (PLMs) with structure adaptations to handle tabular data. For example, TaBERT~\cite{yin-etal-2020-tabert} incorporates vertical self-attention to unify textual and tabular representations based on BERT~\cite{devlin-etal-2019-bert}. TUTA~\cite{wang2021tuta} employs a hierarchical tree-based attention with column and table segment masking to learn table representations. TAPAS~\cite{herzig-etal-2020-tapas} extends BERT by adding row, column, and rank embeddings, along with table cell masking. TAPEX~\cite{liu2022tapex} pre-trains BART~\cite{lewis-etal-2020-bart} using a large synthetic dataset derived from \texttt{WikiTableQuestions}.  
OmniTab~\cite{jiang-etal-2022-omnitab} pre-trains BART on both real and synthetic data, i.e.,  SQL queries from the \texttt{Spider} dataset and synthetic sentences generated from SQL queries via an SQL-to-NL module.  CABINET~\cite{patnaik2024cabinet} proposes a relevance scorer and a relevant cell predictor to assign weights to table cells.
These single-stage PLM-based models struggle with large and noisy inputs due to their limited context windows and reasoning capabilities.

\underline{LLM prompting.} Recent studies leverage LLMs via prompting. Mix-SC \cite{liu-etal-2024-rethinking} issues multiple prompts using LLMs either as a Python agent or with Chain-of-Thought (CoT) prompting~\cite{wei2022chain}. It adopts a self-consistency voting strategy to select the most frequent answer of the prompted LLM. 
Binder~\cite{cheng2023binding} iteratively refines an LLM-generated program and executes it to obtain answers, while TabLaP~\cite{wang2025accurate} follows a similar approach but focuses on numerical questions. These models still ingest the full input table. Their performance degrades on large tables, as shown earlier. 

\textbf{Decompose-then-Reason Models} 
Decompose-then-reason models reduce the size of the input table first, such that the reasoning process (e.g., with an LLM) can focus on more targeted sub-tables. Studies taking this approach can be categorized by their table decomposition strategies.  

\underline{SQL execution-as-decomposer.} A natural choice to decompose tables (especially fully structured ones) is to run SQL queries. NL-to-SQL-based methods translate natural language questions into SQL queries that are directly executed to obtain the final answer. Such methods can also be seen as a type of single-stage method, as no reasoning is needed afterwards.  
The SQL queries can be generated with either a trained model or LLMs via in-context learning. For example, Seq2SQL~\cite{zhong2017seq2sql} trains a seq2seq model to generate SQL queries. DIN-SQL~\cite{pourreza2023} decomposes  SQL generation into sub-tasks to improve accuracy. LEVER~\cite{ni2023lever} generates multiple candidate answers via SQL execution and trains a verifier to select the most accurate one. SynTQA~\cite{ZhangLZ24} chooses answers from SQL execution and an end-to-end TableQA module. 

There are also studies that run SQL queries to extract sub-tables for LLM reasoning. OpenTab~\cite{Kong0SSLFRK24} considers an open-domain setup where tables relevant to the input question are first retrieved using BM25~\cite{RobertsonZ09}. Then, it runs SQL queries generated via few-shot prompting to obtain sub-tables. TabSQLify~\cite{NahidR24} runs SQL queries generated based on the question and table headers to retrieve a sub-table.

\underline{Program-as-decomposer.} TableRAG~\cite{ChenME00CFLLP24} uses LLM to iteratively generate Python programs based on the question, table headers, and reasoning contexts. The programs are executed to retrieve key cells and columns from the original table. An LLM reasons over the retrieved data to generate the final answer. 

\emph{The models above rely heavily on the correctness and executability of SQL queries and programs, which could easily fail due to the generative nature of the queries and programs and the complex structure of the tables.} 

\underline{Model-as-decomposer.} 
Another series of studies uses PLMs or LLMs as the table decomposer. Naively, one can put the full input table and question into an LLM to locate the table rows and columns (or their indexes) relevant to the question. Learn-to-Reduce~\cite{lee2024learning} uses a column selector and a row selector, which are fine-tuned language models. These methods perform one-shot sub-table selection without iterative refinement or verification, making them vulnerable to failures of the selectors. Other studies further introduce \emph{multi-stage decomposition-and-reasoning}. Chain-of-Table~\cite{wang2024chainoftable} extends the CoT framework by progressively extracting a sub-table at each reasoning step. DATER~\cite{ye2023large} follows a similar idea, decomposing both the original table and the input question into sub-tables and sub-questions, which are then used to answer questions. These methods also use models (e.g., LLMs) as the decomposer. They still feed the models with the full input tables. This challenges the models' semantic understanding capabilities over a long input context. 


\begin{figure*}[t]
     \centering
    \includegraphics[width = 1\linewidth]{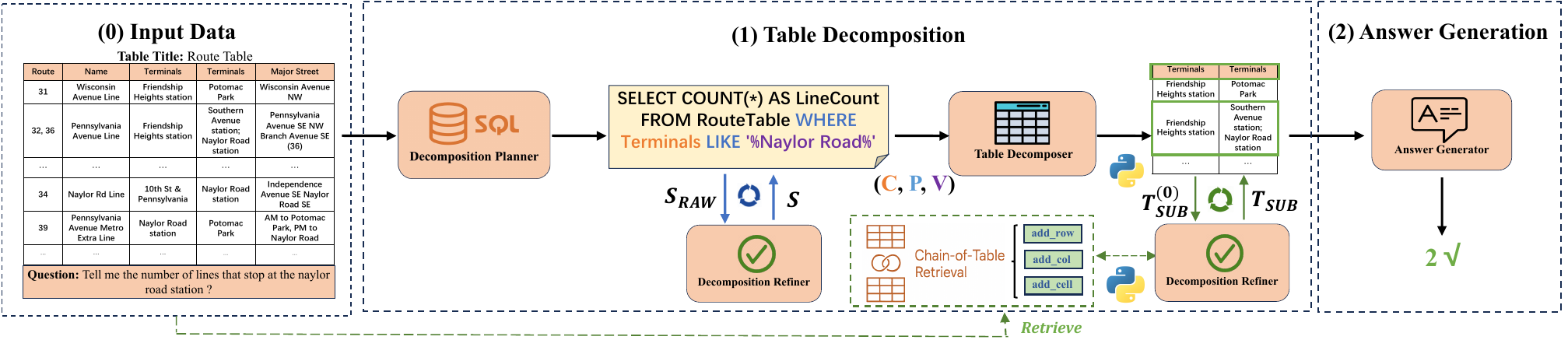}
     \caption{Overview of \model. The model contains two stages: (i)~\textit{Table Decomposition}, where an SQL query is generated and refined to guide sub-table extraction based on decomposition triples (Column Names, Conditions, Values), highlighted in orange, blue, and purple, respectively; and (ii)~\textit{Answer Generation}, where the question and the refined sub-table are used to prompt an LLM for the final answer.}
     \label{fig:model_overview}
\end{figure*}

Our model \model\ follows the decompose-then-reasoning paradigm. Instead of applying either SQL queries or LLMs directly on the full input tables, we use both to generate a decomposition plan, hence bypassing the limitations of both approaches and enhancing the quality of the decomposition outcomes. 


\section{Problem Formulation}
Given a table \( T \) and a question \( Q \) about the informtion in \( T \), 
the TableQA task aims to develop a model capable of generating an accurate answer \( A \) for \( Q \). Specifically, each question \( Q = (q_1, q_2, \dots, q_{|Q|}) \) is expressed in natural language, where $q_i \in Q$ ($i \in [1, |Q|]$) represents a token (e.g., a word). The table \( T \) is also represented as a token sequence in natural language, where individual cells are separated by special characters such as `$\vert$', and rows are delineated by newlines. 

For TableQA over large tables which are of more than 4,096 tokens (the context window size limit of most TableQA PLMs), the aim becomes to identify a sub-table $T^* \subseteq T$ that allows a reasoning model $R(\cdot,\cdot)$ to answer $Q$ correctly while minimizing the length of $T^*$. 
Formally: 
\begin{equation}
T^* = \arg\min_{T' \subseteq T} |T'| \text{ subject to } {R(T', Q) = A}.
\label{eq:large_qa_problem_opt}
\end{equation}
Here $|\cdot|$ denotes the length of a table. 

In practice, it is rather difficult, if not impossible, to obtain the minimum $T^*$, otherwise $Q$ may have been answered already (e.g., when $T^* = A$). We reduce the goal to obtaining a size-reduced sub-table $\hat{T}^*$ that maximizes the probability for  $R(\cdot,\cdot)$, which is an imperfect model and could struggle with large table input, to answer $Q$ correctly, 

\begin{equation}
\hat{T}^* = \arg\max_{T' \subseteq T} \mathbb{P}(R(T', Q) = A).
\label{eq:large_qa_problem}
\end{equation}

Based on $\hat{T}^*$, we run $R(\hat{T}^*, Q)$, which could be a PLM or LLM, to generate the TableQA answer $\hat{A}$.



\section{Methodology}
\model\ runs in two stages: (i)~\emph{Table Decomposition} extracts a compact and relevant sub-table $T_{\text{SUB}} \subset T$; (ii)~\emph{Answer Generation}  predicts the final answer $\hat{A}$ based on $T_{\text{SUB}}$ and $Q$, as outlined in Algorithm~\ref{alg:model} and Figure~\ref{fig:model_overview}.



\begin{algorithm}[t]
\caption{TableQA with \model{}}
\label{alg:model}
\small
\begin{algorithmic}[1]
\REQUIRE Table $T$, Question $Q$; Models: \smodel\ $M_{\text{P}}$, \vmodel\ $M_{\text{V}}$, \dmodel\ $M_{\text{D}}$, \amodel\ $M_{\text{A}}$
\ENSURE Answer $A$

\STATE \textbf{Stage I: Table Decomposition}
\STATE $S_{\text{RAW}} \leftarrow M_{\text{P}}(Q, \text{header}(T))$
\STATE $S \leftarrow M_{\text{V}}^{(1)}(S_{\text{RAW}}, Q, \text{header}(T))$ \hfill // Pre-decomposition refinement
\STATE $(C, P, V) \leftarrow M_{\text{D}}.\text{Parse}(S)$
\STATE $C^* \leftarrow \text{LCS-Correct}(C, \text{header}(T))$
\STATE $T_{\text{SUB}}^{(0)} \leftarrow M_{\text{D}}.\text{Filter}(T, C^*, P, V)$
\STATE $\hat{A} \leftarrow M_{\text{V}}^{(2)}(Q, \text{header}(T), T_{\text{SUB}}^{(0)})$ \hfill // Post-decomposition refinement

\IF{$\hat{A} \ne \emptyset$}
    \RETURN $\hat{A}$
\ELSE
    \STATE $\mathcal{A}_{\text{CoTR}} \leftarrow M_{\text{V}}^{(2)}.\text{Actions}(Q, T_{\text{SUB}}^{(0)}, T)$
    \STATE $T_{\text{SUB}} \leftarrow T_{\text{SUB}}^{(0)} + \text{Execute}(\mathcal{A}_{\text{CoTR}}, T)$
\ENDIF

\STATE \textbf{Stage II: Answer Generation}
\STATE $\hat{A} \leftarrow M_{\text{A}}(Q, T_{\text{SUB}})$
\RETURN $\hat{A}$
\end{algorithmic}
\end{algorithm}

\subsection{Stage I: Table Decomposition}
\subsubsection{SQL Generation via Few-shot Prompting} We first use a \smodel\ to generate a raw SQL query $S_{\text{RAW}}$ based on question $Q$ and the header of table $T$. \smodel\ is prompted with top-$K$ in-context examples retrieved from an NL-to-SQL dataset \texttt{SQUALL}~\cite{ShiZBDL20} (see prompt template in Appendix A.1). We compute the cosine similarity between the embeddings of $Q$ and all $N$ questions in \texttt{SQUALL}, denoted as $D = \{d_1, d_2, \ldots, d_N\}$, using the Sentence Transformer~\cite{Song0QLL20}:
\begin{equation}
\mathbf{e_Q} = f(Q), \quad \mathbf{E_D} = \{ f(d_i) \}_{i=1}^{N}\,,
\end{equation}
where $f(\cdot)$ is an embedding model, $\mathbf{e_Q} \in \mathbb{R}^{1 \times m}$, $\mathbf{E_D} \in \mathbb{R}^{N \times m}$, and $m$ denotes embedding  dimensionality. 

We select the top $K$ examples based on cosine similarity:
\begin{equation}
\mathbf{Z} = \operatorname{argsort}(\frac{\mathbf{e_Q} \mathbf{E_D}^T}{\|\mathbf{e_Q}\| \cdot \|\mathbf{E_D}\|})\,,
\end{equation}
\begin{equation}
\operatorname{TopK}(D) = \{ d_{i_j} \mid i_j \in \mathbf{Z}\}, j\in [N-K+1\colon N]\,,
\end{equation}
where $i \in [1, |D|]$, and $K = 5$ by default.

\paragraph{SQL Verification and Refinement (pre-decomposition)} The raw SQL query $S_{\text{RAW}}$ is verified by the \vmodel, which prompts an LLM with $S_{\text{RAW}}$, the full table header, and question $Q$, to check if $S_{\text{RAW}}$ leads to an answer of $Q$. If not, the \vmodel\ revises $S_{\text{RAW}}$ into a refined SQL query $S$. This process could repeat, although we do not repeat by default,  for cost efficiency. See the prompt template in Appendices A.2 and A.3.

\paragraph{Sub-table Construction} The \dmodel\ parses $S$ using regex (detailed in Appendix B.1) to extract a triple $(C, P, V)$ representing column names, conditions (e.g., filtering clauses), and values (e.g., values in the conditions). We choose regex over an LLM-based parser because it achieves higher accuracy empirically (see Appendix B.2). To mitigate potential hallucinations in $C$ (i.e., resulting from LLM SQL generation), we compute the longest common subsequence (LCS) ratio between each extracted column name and the column names of $T$. A column $c_i \in C$ is replaced with the matched column name from $T$ if the best LCS score exceeds 0.2, or discarded otherwise.

We construct the initial sub-table $T_{\text{SUB}}^{(0)}$ using the corrected tripe $(C^{\ast}, P, V)$ by applying column projection and row filtering over the DataFrame form of $T$. Each SQL-style condition $p_i \in P$ is mapped to a corresponding DataFrame filtering operation (see Appendix~B.3). For string-based filters, we adopt a fuzzy matching strategy where strict equality conditions (``\texttt{=}'' in SQL) are relaxed to substring matching (``\texttt{LIKE}'' in SQL), implemented using Python's \texttt{in} operator. If filtering fails, all rows in the target column are retained.

\paragraph{CoTR: Sub-table Refinement (post-decomposition)} 
The \vmodel\ also verifies the completeness of $T_{\text{SUB}}^{(0)}$. If $T_{\text{SUB}}^{(0)}$ is insufficient for answering  $Q$, we present a Chain-of-Table-Retrieval (CoTR) mechanism inspired by the \textit{Chain-of-Table} paradigm~\cite{wang2024chainoftable}. 
CoTR uses an LLM to iteratively generate retrieval actions, such as to retrieve more rows, columns, or cells, to incrementally augment $T_{\text{SUB}}^{(0)}$. These actions are implemented as executable Python instructions over $T$, leading to a more complete sub-table $T_{\text{SUB}}$ (the prompts used are in Appendix A.4). 

\subsection{Stage II: Answer Generation}\label{sec:ans_gen} 
At this stage, we leverage the strong semantic-parsing and reasoning capabilities of LLM to perform \emph{end-to-end} TableQA. If the \vmodel\ in the previous stage has already produced a complete answer, it is returned directly. Otherwise, we feed the refined sub-table $T_{\text{SUB}}$ and question $Q$ (see prompt template in Appendix A.5) into the \amodel\ (an LLM) to generate the final answer $A$.

\section{Dataset Construction}\label{sec:datasets}
Prior work typically tests their models on a small subset of large tables from existing TableQA datasets (e.g., \texttt{WikiTableQuestions}), where the proportion of such tables is limited. Other efforts use large tables from NL-to-SQL datasets such as \texttt{Spider}. The QA pairs from such datasets are tailored for SQL generation and \emph{not} for queries over semi-structured tables. 

\begin{figure}[t]
     \centering
     \includegraphics[width = 1\linewidth]{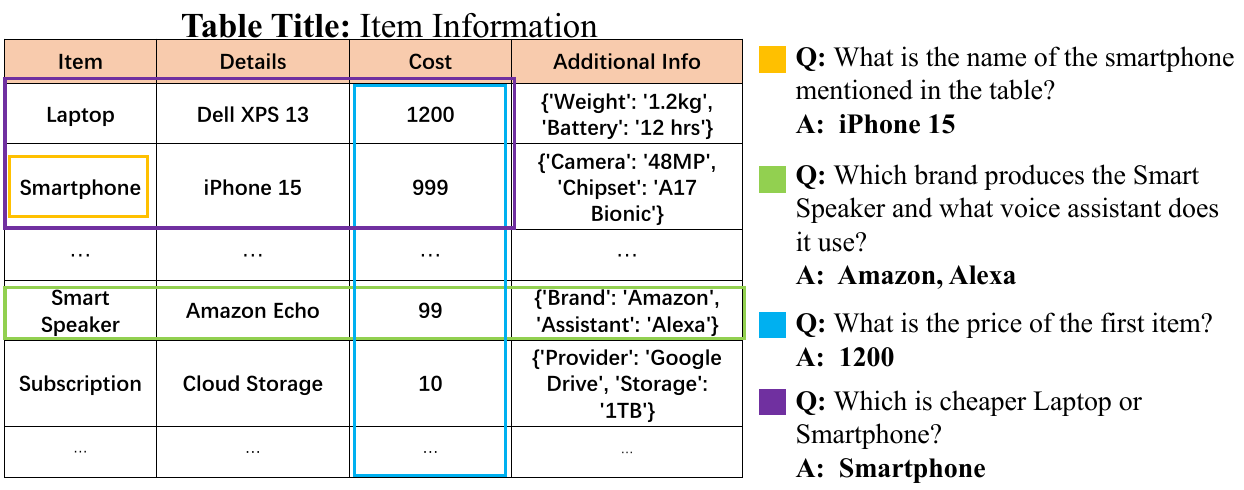}
     \caption{Generated QA pair examples.}
     \label{fig:qa_gen}
\end{figure}

To address these issues, we construct two new large-table TableQA datasets: \texttt{\ldataset} and \texttt{\sdataset}, which contain QA pairs automatically generated through a self-adaptive QA generation pipeline using \texttt{GPT-4o mini}~\cite{openai2024gpt4omini}, with minimal human annotation and correction. Our approach achieves over 70\% QA pair acceptance rate (see Appendix C.1), allowing for scalable and cost-efficient dataset construction.  Figure~\ref{fig:qa_gen} shows examples of generated QA pairs.

\textbf{Self-adaptive QA pair generation.} We observe that directly prompting an LLM with a selected cell or sub-table as the ground-truth answer to generate a corresponding question often leads to low-quality QA pairs (acceptance rate $< 50\%$ based on human evaluation). To overcome this issue, we propose a \emph{self-adaptive} QA pair generation method, where the LLM generates an answer -- derived either from a specific cell or an answer field (i.e., row, column, or sub-table) -- and then constructs the corresponding question. The generated QA pairs are further manually reviewed and post-processed: questions deemed unreasonable are revised, and incorrect answers are corrected to ensure the final QA pair is valid. This method significantly improves QA pair quality, with fewer than 30\% of the QA pairs requiring human correction (see details in Appendix C.1). 

We propose three answer field selection strategies: cell-based, column/row-based, and sub-table-based, which prompt the LLM to randomly select a cell, a column/row, and a sub-table (containing multiple rows and columns) as the ground-truth answer, respectively, and to generate a QA pair based on the answer (see prompts in Appendix C.2).


\texttt{\sdataset} is construed by extracting all the large tables (with more than 4,096 tokens) from the \texttt{Spider} dataset. 
\texttt{Spider}'s original QA pairs are designed for NL-to-SQL tasks, which tend to align with rigid logical forms and thus lack the diversity and complexity required for TableQA testing. To address this gap, we regenerate QA pairs using our LLM-based pipeline above. Importantly, we include both SQL-executable questions and non-SQL-executable questions (detailed in Appendices C.4 and C.5; see prompts and analysis in Appendix C.6), enabling more challenging reasoning beyond rigid logical forms.

\texttt{\ldataset} first expands the small tables (with 4,096 tokens or fewer) in \texttt{Spider} via LLM-based row and column augmentation, while maintaining structure consistency (see Appendix C.3). The expanded tables are then used with the same QA generation as above to produce diverse questions. 


\section{Experiments}
\subsection{Experimental Settings} \label{sec:experiment} 
\paragraph{Datasets} 
We evaluate \model\ on five datasets: the two new large-table benchmarks constructed by us \texttt{\sdataset} and \texttt{\ldataset}, and three 
public datasets: \texttt{WikiTableQuestions} (\texttt{\wdataset}), \texttt{HybridQA} (\texttt{\hdataset}) and \texttt{TabFact}.

\texttt{\wdataset} is derived from Wikipedia tables. To focus on large tables, we extract all tables with more than 4,096 tokens from the \texttt{\wdataset} dataset to form a large-table subset, denoted as \textbf{\wdataset$^{L}$}. We also evaluate on the full  \texttt{\wdataset} dataset to assess model generalizability across different table sizes.

\texttt{\hdataset} combines semi-structured tables with texts from Wikipedia. 
To maintain consistency in prompt formatting, we include the texts directly in the corresponding table cells. Then, we take all the tables with more than 4,096 tokens to construct a large-table subset, denoted as \textbf{\hdataset$^{L}$}.

\texttt{\tdataset} is a table-based fact verification benchmark. We use it to assess  \model’s capability to handle different TableQA tasks beyond standard question answering.

Table~\ref{tab:dataset-features} summarizes the five datasets.

\begin{table}[t]
\scriptsize
\centering
\setlength{\tabcolsep}{0.7pt}
\begin{tabular}{ccccccccc}
\toprule
\multicolumn{1}{c}{\multirow{2}{*}{\textbf{Dataset}}} &  \textbf{\#  Cols.} & \textbf{\#  Rows} & \textbf{\# Toks.} &  \textbf{\# Toks.} & \textbf{\# Toks.} & \multicolumn{3}{c}{\textbf{\#  QA Pairs}} \\
\multicolumn{1}{c}{}  & \textbf{/ Table} &  \textbf{/ Table} & \textbf{/ Table} &  \textbf{/ Cell} &\textbf{/ Answer} &\textbf{Train} &\textbf{Validation} &\textbf{Test} \\     \midrule
\texttt{\tdataset} & 6.3 & 13.5 & 317.9 & 3.7 & 1.0 & 92,283 & 12,792 & 2,024 \\
\midrule
\texttt{\wdataset}    &   6.4  & 25.4  & 662.6 & 4.1 & 1.7 &   11,321  & 2,831 &  4,344 \\ 
\texttt{WTQ$^{L}$}   &   6.8 & 232.8   &  6,860.8  & 4.3 & 2.9 &   191  & 33 &  81\\ 
\midrule
\texttt{HybQA$^{L}$} & 4.6 & 16.9 & 9,035.7 & \textbf{116.2} & \textbf{4.4}  & 48,541 & 2,694 & 2,630 \\
\midrule
\texttt{\sdataset} & \textbf{11.0} & \textbf{733.8} & 9,786.2 & 1.2 & 2.8 & 1,324 & 239 & 1,110\\
\midrule
\texttt{\ldataset} & 10.7 & 519.6 & \textbf{18,291.5} & 3.3 & 3.7 & 226 & 46 & 169 \\

\bottomrule
\end{tabular}
\caption{Dataset statistics (all values are average values). `Cols.' means `Columns'; 
`Toks.' means `Tokens'; `$*^{L}$' denotes a dataset adapted to include only the large tables.}
\label{tab:dataset-features}
\end{table}

\paragraph{Model Input Preparation} 
Following the literature~\cite{liu-etal-2024-rethinking, wang2024chainoftable, wang2025accurate}, tables are linearized into sequences by flattening their structure and incorporating them into prompts for LLMs. 
Table cells are delineated using  `$\vert$'s, 
while rows are separated by line breaks. 
The questions are directly included in the prompts. 


For our \model\ model, the {\smodel} takes the question, the table header, and in-context examples to generate an SQL query.
The {\dmodel} takes the generated SQL query and the original table to extract a raw sub-table.
The {\vmodel}, at pre-decomposition, takes the raw SQL query, the question, and the table header to generate a refined SQL query; at post-decomposition, it takes the question, the table header, and the initial sub-table to generate a refined sub-table. 
The {\amodel} takes the question and the refined sub-table as input, forms a compact prompt for the LLM to generate the final TableQA answer.

\paragraph{Competitors}
We compare with models of three categories: (i)~NL-to-SQL based: DIN-SQL~\cite{pourreza2023}; (ii)~Fine-tuned PLM-based: TAPEX-Large~\cite{liu2022tapex}, OmniTab-Large~\cite{jiang-etal-2022-omnitab}, and CABINET~\cite{patnaik2024cabinet} (SOTA); (iii) LLM-based: TableRAG~\cite{ChenME00CFLLP24}, TabSQLify~\cite{NahidR24}, DATER~\cite{ye2023large}, and End-to-End QA (with full Tables) ~\cite{openai2024gpt4omini, abs-2312-11805} (SOTA). 

\paragraph{Implementation Details}
We use \texttt{GPT-4o mini} (default and shared by all LLM-based models) and \texttt{Gemini~2.0} as the LLM backbones. All baseline models use their default context window sizes if not specified. We use \texttt{all-mpnet-base-v2} from the Sentence Transformers library~\cite{ReimersG19} as the encoder for SQL example selection. For the study on token length, we use the \texttt{cl100k\_base} tokenizer from \texttt{tiktoken}. All experiments are run with an NVIDIA A100 80 GB GPU.

\paragraph{Evaluation Metric} We report the performance based on the exact-match \textbf{Accuracy}, following the official evaluator from~\citet{pasupat-liang-2015-compositional}. 

\begin{table}[ht!]
\centering
\small
\setlength{\tabcolsep}{1pt}
\begin{tabular}{lllcccc}
\toprule
\multicolumn{2}{c}{\textbf{Model}} & \texttt{\wdataset$^{L}$} & \texttt{\hdataset$^{L}$} & \texttt{\sdataset} & \texttt{\ldataset}\\
\midrule
{\multirow{1}{*}{NL-to-SQL}} & DIN-SQL &  $41.98$ & $0.27 $& $49.82$ & \underline{$69.82$}\\
\midrule
{\multirow{3}{*}{Fine-tuned}} & OmniTab-Large   &  $29.63$  & $7.83$ & $20.95$ & $13.02$ \\
& TAPEX-Large     &  $30.86$  & $6.46$ & $17.09$ & $10.65$ \\
& $\dag$ CABINET     &  $43.21$  & $8.72$ & $21.47$ & $13.61$ \\
\midrule
{\multirow{16}{*}{LLM-based}} 
& \multicolumn{5}{c}{Base: \texttt{GPT-4o mini}} \\  
& $\dag$ TableRAG & $30.86$ & $9.89$ & $27.12$ & $36.69$ \\
& $\dag$ TabSQLify & $32.10$ & $21.60$ & $22.97$ & $15.38$ \\
& $\dag$ DATER   &  $35.80$  & $26.81$ & $44.95$ & $53.25$ \\
& End-to-End QA & \underline{$48.15$} &  \underline{$30.95$} & \underline{$57.75$} & $64.50$ \\
& \textbf{\model~(ours)} &  $\mathbf{61.73}$ & $\mathbf{36.01}$ & $\mathbf{71.35}$ & $\mathbf{76.33}$ \\

& $\Delta$ & $\uparrow28.20$ & $\uparrow16.35$ & $\uparrow23.55$ & $\uparrow9.32$\\
\cmidrule(lr){2-6} 
& \multicolumn{5}{c}{Base: \texttt{Gemini 2.0}} \\  
& $\dag$ TableRAG & $41.98$ & $15.02$ & $34.23$ & $38.46$ \\
& $\dag$ TabSQLify & $40.74$ & $19.90$ & $25.67$ & $21.89$ \\
& $\dag$ DATER   &  $39.51$  & $18.78$ & $47.03$ & $52.07$ \\
& End-to-End QA & \underline{$60.49$} &  \underline{$36.24$} & \underline{$65.41$} & \underline{$71.01$} \\
& \textbf{\model~(ours)} &  $\mathbf{70.37}$ & $\mathbf{38.02}$ & $\mathbf{69.64}$ & $\mathbf{77.51}$ \\
& $\Delta$ & $\uparrow16.33$ & $\uparrow4.91$ & $\uparrow6.47$ & $\uparrow9.15$\\
\bottomrule
\end{tabular}
\caption{Overall performance results. 
The best results are in \textbf{boldface}, while the best baseline results are \underline{underlined}; $\Delta$ (\%) denotes the performance gain of \model\ compared with the best baseline results. $\dag$ denotes baselines with table decomposition. The other baselines take full tables as input.} 
\label{tab:model_results}
\end{table}

\subsection{Results and Analysis}
\paragraph{Overall Results} Table~\ref{tab:model_results} reports the overall performance results. Our model \model\ consistently outperforms all competitors across all four datasets using two different backbone LLMs. This confirms the effectiveness of  \model\ on dealing with large tables.

Among the datasets, \texttt{\hdataset}$^L$ is the most challenging due to its mix with lengthy textual contents in cells. This invalidates the NL-to-SQL-based model DIN-SQL. Although the End-to-End QA model is relatively stable across datasets due to the strong textual understanding capability of the backbone LLMs, its accuracy degrades as the table length increases, as shown in Figure~\ref{fig:token_acc}. 
In contrast, \model\ is just as robust while its accuracy is higher, benefiting from its table decomposition process to reduce the table length and simplify LLM reasoning (comparison shown in Appendix~B.4). 



\begin{figure}[t]
     \centering
     \includegraphics[width = 1 \linewidth]{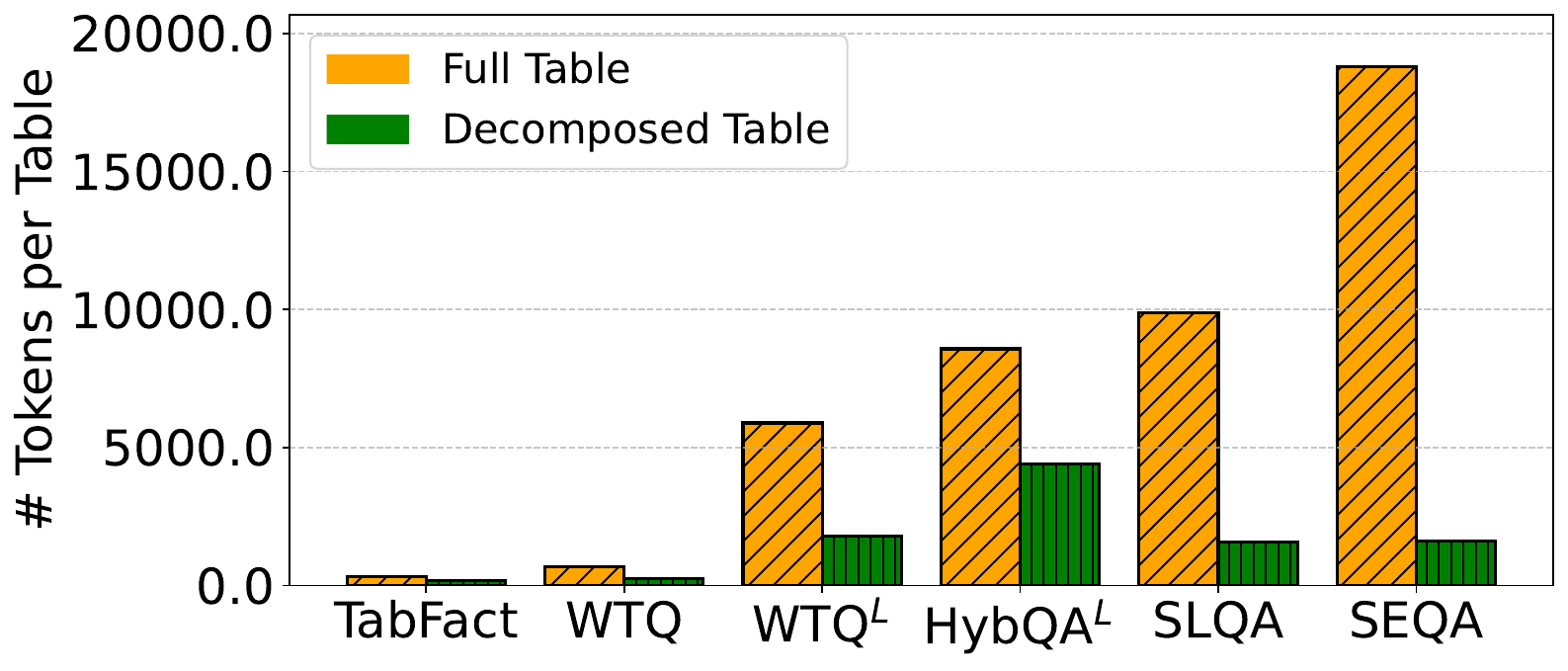}
     \caption{
     Comparison of the lengths of tables input to the LLM reasoning module with and without running \dmodel\ of \model\ on the test sets of the datasets.}
     \label{fig:token_eff}
\end{figure}

\paragraph{Effectiveness of the \dmodel} As shown in Figure~\ref{fig:token_eff}, our \dmodel\ effectively reduces the length of the table fed to the LLM in the \amodel\ by $40.0\%$, $63.5\%$, $69.8\%$, $48.5\%$, $84.0\%$, and $91.4\%$ on the six datasets, respectively. This substantial reduction enables the \amodel\ to focus on important sub-tables while enhancing data efficiency.

\begin{figure}[t]
     \centering
     \includegraphics[width = 1 \linewidth]{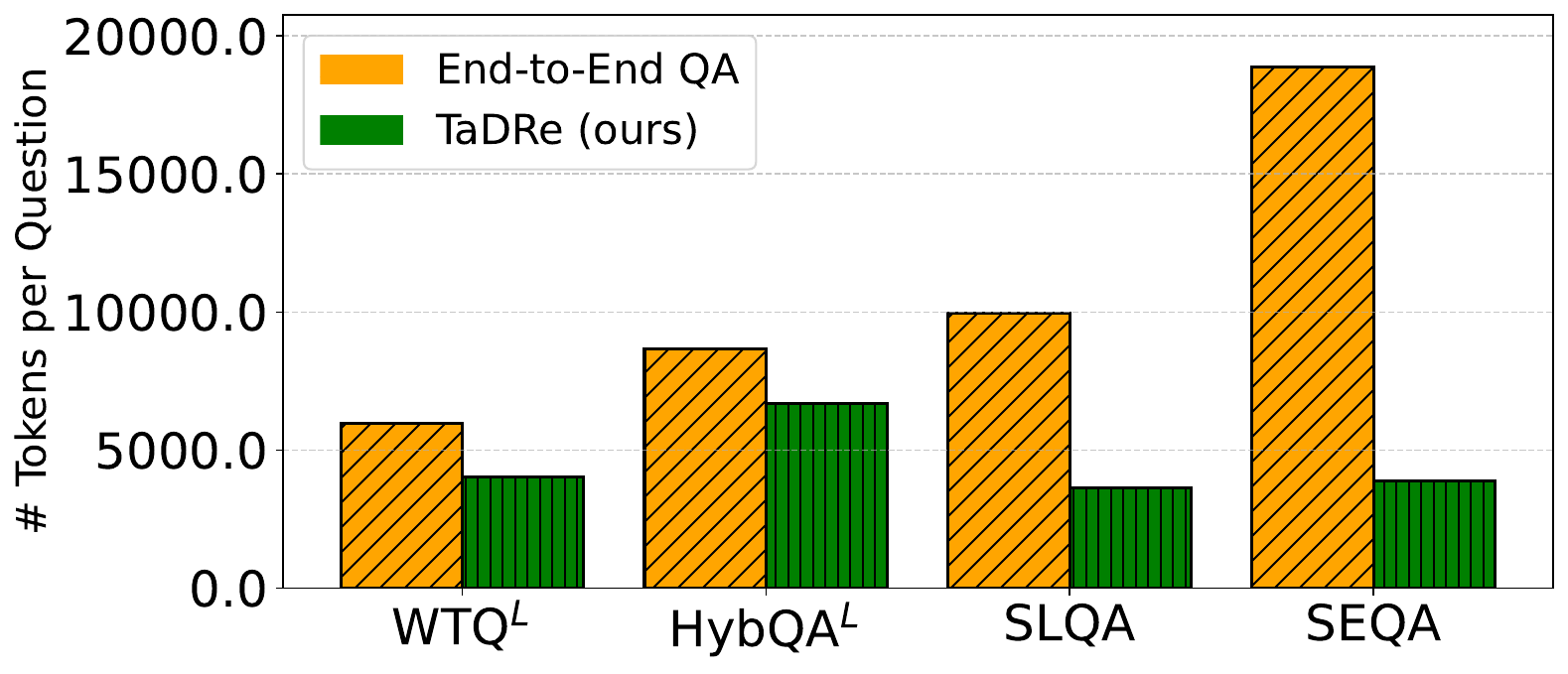}
     \caption{
     Comparison of the overall number of LLM input tokens per question between \model\ and the End-to-End QA model.}
     \label{fig:input_tokens}
\end{figure}

As \model\ also uses LLMs in its other modules, 
Figure~\ref{fig:input_tokens} further compares the total number of input tokens per question for all LLMs in \model\  with that of the End-to-End QA. \model\  reduces the overall LLM input size by $32.6\%$, $22.6\%$, $63.5\%$, and $79.4\%$ on the four large-table benchmarks, \texttt{WTQ$^{L}$}, \texttt{HybQA$^{L}$}, \texttt{\sdataset}, and \texttt{\ldataset}, respectively.

\begin{table}[t]
\centering
\small
\setlength{\tabcolsep}{3pt}
\begin{tabular}{lcccc}
\toprule
\textbf{Model} & \texttt{\wdataset$^L$} & \texttt{\hdataset$^L$} & \texttt{\sdataset} & \texttt{\ldataset} \\
\midrule
TAPEX-Large & $30.86$ & $6.46$ & $17.09$ & $10.63$ \\
w/ LLM-Decomposer   & $25.93$ & $5.89$ & $17.66$ & $8.28$ \\
\textbf{w/ \model-Decomposer}        & $\textbf{39.51}$ & $\textbf{11.60}$ & $\textbf{40.27}$ & $\textbf{42.60}$ \\
\cmidrule{2-5}
OmniTab-Large & $29.63$ & $7.83$ & $20.95$ & $13.02$ \\
w/ LLM-Decomposer     & $24.61$ & $6.43$ & $20.10$ & $11.24$ \\
\textbf{w/ \model-Decomposer}            & $\textbf{46.92}$ & $\textbf{10.80}$ & $\textbf{54.23}$ & $\textbf{50.89}$ \\
\midrule
\texttt{Qwen2.5-3B-Instruct} & $25.93$ & $19.58$ & $27.57$ & $26.04$ \\
w/ LLM-Decomposer & $25.93$ & $14.07$ & $21.35$ & $11.83$ \\
\textbf{w/ \model-Decomposer}        & $\textbf{39.51}$ & $\textbf{23.19}$ & $\textbf{49.46}$ & $\textbf{59.17}$ \\
\cmidrule{2-5}
\texttt{Qwen2.5-32B-Instruct} & $41.98$ & $34.98$ & $54.05$ & $53.25$ \\
w/ LLM-Decomposer   & $32.10$ & $14.20$ & $32.34$ & $19.51$ \\
\textbf{w/ \model-Decomposer}          & $\textbf{58.02}$ & $\textbf{36.12}$ & $\textbf{66.31}$ & $\textbf{72.78}$ \\
\bottomrule
\end{tabular}
\caption{Performance gains from applying our \dmodel\ to PLM-based and ``small'' LLM-based models.} 
\label{tab:enhancement}
\end{table}

The substantial reduction allows the sub-tables to fit the input window constraints of the fine-tuned PLM-based models and ``small'' LLMs, creating opportunities to achieve high accuracy with such models.  
Table~\ref{tab:enhancement} verifies this by powering PLM-based models TAPEX-Large and OmniTab-Large and ``small'' LLMs \texttt{Qwen2.5-3B-Instruct} and \texttt{Qwen2.5-32B-Instruct}~\cite{qwen2025} with our \dmodel.  We observe consistent and substantial accuracy gains with our \dmodel. 
Notably, \texttt{Qwen2.5-32B-Instruct} with our \dmodel\ matches or even surpasses the performance of \texttt{GPT-4o mini}- and \texttt{Gemini 2.0}-based End-to-End QA models (72.78 on \texttt{SEQA} vs. 64.50 and 71.01 as shown in Table~\ref{tab:model_results}).

We compare with an LLM-Decomposer, i.e., directly prompting \texttt{GPT-4o mini} to return the indexes of row and column relevant to input questions (see the prompt in Appendix D.1). We see that this decomposer leads to reduced accuracy, because the LLM struggles to reason effectively over large tables in full. 

\paragraph{Model Applicability} Table~\ref{tab:model_results} has reported the applicability of \model\ to different backbone LLMs  on large tables. Here, we further evaluate \model's applicability on the full \texttt{\wdataset} dataset containing smaller tables and the \texttt{\tdataset} dataset for the Table Verification task.

Table~\ref{tab:standard} shows that \model\ also outperforms the baselines on the full \texttt{\wdataset} dataset -- see Appendix B.2 for a detailed breakdown of accuracy on tables of different lengths. 
On the table verification dataset \texttt{\tdataset}, \model\ is slightly less accurate than End-to-End \texttt{GPT-4o mini}. This is likely due to the small tables in \texttt{\tdataset} (cf. Table~\ref{tab:dataset-features}), which are easy for \texttt{GPT-4o mini} while table decomposition poses risks in omitting question-relevant information. 

\begin{table}[t]
\centering 
\setlength{\tabcolsep}{2pt}
\small
\begin{tabular}{llcc}
\toprule
\multicolumn{2}{c}{\textbf{Model}} & \texttt{\wdataset} & \texttt{\tdataset} \\
\toprule
NL-to-SQL & DIN-SQL & $53.11$ & $69.17$\\
\midrule
{\multirow{3}{*}{Fine-tuned}} & TAPAS-Large & $48.80$ & $78.46$\\
 & TAPEX-Large & $57.50$ & $75.30$\\
 & OmniTab-Large & $61.21$ & $85.15$\\
\midrule
{\multirow{8}{*}{LLM-based}} 
& \texttt{Qwen2.5-3B-Instruct} & $37.25$ & $66.95$\\ 
& $\ast$\texttt{Qwen2.5-3B-Instruct} & $38.28$ & $73.37$\\ 
& End-to-End QA & $50.87$ & $\textbf{86.46}$\\
& TableRAG & $54.10$ & $69.57$\\
& TabSQLify & $56.49$ & $70.55$\\
& DATER & $62.15$ & $82.36$\\
& \textbf{\model~(ours)} &  $\textbf{64.27}$ & $83.89$\\
\toprule
\end{tabular}
\caption{Results on the full \texttt{\wdataset} dataset and the \texttt{\tdataset} dataset. $\ast$ represents model with our \dmodel.}
\label{tab:standard}
\end{table}

\begin{table}[t]
\centering
\small
\setlength{\tabcolsep}{2pt}
\begin{tabular}{lcccc}
\toprule
\multicolumn{1}{c}{\textbf{Model}}
  & \texttt{\wdataset$^{L}$} & \texttt{\hdataset$^{L}$} & \texttt{\sdataset} & \texttt{\ldataset} \\
\midrule
DP SQL execution & $23.46$ & $0.42$  & $12.61$ & $26.04$ \\
 \midrule
\model-w/o-DP+TD  & {$48.15$} & {$30.95$} & {$57.75$} & {$64.50$}\\
 \midrule \model-w/o-DR (Pre) & {$56.79$} & {$28.40$} & {$69.01$} & {$72.25$}\\
 \midrule
 \model-w/o-DR (Post) & $58.02$ & $31.83$ & $70.10$ & $75.15$ \\
  \midrule
 \textbf{\model} &  $\mathbf{61.73}$ & $\mathbf{36.01}$ & $\mathbf{71.35}$ & $\mathbf{76.33}$ \\

\bottomrule
\end{tabular}
\caption{Ablation study results. `DP' denotes the \smodel; `TD' denotes the \dmodel; `DR' denotes the \vmodel.}
\label{tab:ablation}
\end{table}

\paragraph{Ablation Study} We run an ablation study with four model variants: (1)~\textbf{DP SQL execution}: Directly executing SQL queries generated by the \smodel; 
(2)~\textbf{\model-w/o-DP+TD}: \model\ without both the \smodel\ and the \dmodel, i.e., only using the \amodel; (3)~\textbf{\model-w/o-DR (pre)}: \model\ without SQL refinement using the \vmodel; 
(4)~\textbf{\model-w/o-DR (post)}: \model\ without sub-table refinement using the \vmodel\ (also excluding the Chain-of-Table-Retrieval). 

As Table~\ref{tab:ablation} shows, directly executing SQL queries (i.e., DP SQL execution) results in substantial accuracy drops, because the SQL queries do not fit the semi-structured tables. Similarly, using only the \amodel\ (\model-w/o-DP+TD) leads to accuracy drops, since LLMs struggle with digesting large tables in full. The \vmodel\ helps improve the generated SQL queries for table decomposition and refines the sub-tables. Removing the module (\model-w/o-DR (pre) and \model-w/o-DR (post)) also hurts model accuracy.

\paragraph{Error Analysis} We count the number of cases where directly executing the generated SQL queries results in execution errors (e.g., BinderException, IndexError, and ParserException). The execution error rates are $64.2\%$, $88.7\%$, $70.6\%$, and $61.5\%$ on \texttt{\wdataset$^L$}, \texttt{\hdataset$^L$}, \texttt{\sdataset}, and \texttt{\ldataset}, respectively, further emphasizing that NL-to-SQL-based solutions struggle on these datasets. See Appendix D.2 for detailed results.  

For the \vmodel, the acceptance rates (i.e., the proportion of cases where $S_{\text{RAW}}$ is considered adequate to guide the \dmodel\ in retrieving an information-complete sub-table without refinement) are $71.6\%$, $54.3\%$, $77.1\%$, and $78.7\%$, respectively. The lower acceptance rate on  \texttt{\hdataset$^L$} ($54.3\%$) suggests that SQL generation struggles more with the hybrid table-text format, thereby requiring more frequent refinement.

We also manually analyze $60$ error cases of \model\ and report results in Appendix D.3. The primary accuracy bottleneck is the backbone LLM's reasoning capability over tabular data. Even when provided with valid sub-tables, the model could still produce incorrect answers.

\paragraph{Other Results} We further include a case study with three TableQA examples that fail existing table decomposition models but not our  \model\ model (Appendix D.4).

\section{Conclusions}
We proposed \model, a TableQA model designed for large, semi-structured tables. \model\ utilizes an LLM-based \smodel\ to generate an SQL query as a decomposition plan, leveraging both structural and semantic information in SQL to guide sub-table construction without direct, error-prone SQL execution. We designed a \dmodel\ that uses pattern-driven extraction to accurately identify table-relevant fields from the SQL query. To further enhance table decomposition quality, the \vmodel\ refines both the SQL query and the sub-tables generated. 
We evaluated \model\ on five datasets. The results show that \model\ consistently outperforms all existing models, establishing new SOTA performance for TableQA over large tables. 

\bibliography{aaai2026}

\clearpage
\appendix

\clearpage

\section{Appendix A. Prompts for Modules of \model}
\subsection{}\label{app:prompts}
The section lists the prompts used by our model components.

\subsection{A.1. Prompt for SQL Generation}
The prompt used for the \smodel\ LLM to generate SQL queries is shown as follows.
\begin{tcolorbox}[colback=gray!10, colframe=black!50, title=Prompt: \smodel, fonttitle=\bfseries]
\small
\ttfamily
/* Some example questions and corresponding SQL queries \\
are provided based on similar problems: */\\
Answer the following: \texttt{[P1]} \\
\texttt{[S1]} \\
Answer the following: \texttt{[P2]} \\
\texttt{[S2]} \\
Answer the following: \texttt{[P3]} \\
\texttt{[S3]} \\
Answer the following: \texttt{[P4]} \\
\texttt{[S4]} \\
Answer the following: \texttt{[P5]} \\
\texttt{[S5]} \\
The target questions: \texttt{[P]} \\
Table columns: \texttt{[Columns]} \\
The corresponding SQL: \\
Note: \\
- Do not add any explanation after the SQL.
\end{tcolorbox}

\subsection{A.2. Prompt for SQL Verification}
The prompt used for the \vmodel\ LLM to verify the generated SQL queries is shown as follows.

\begin{tcolorbox}[colback=gray!10, colframe=black!50, title=Prompt: SQL Verification, fonttitle=\bfseries]
\ttfamily
\small
Please determine whether the sub-table obtained using raw SQL is sufficient to answer the given question. \\
The question is: \texttt{[Question]} \\
Raw Sub-table Columns: \texttt{[Columns]} \\
Raw Sub-table Content: \texttt{[TAB\_CONTENT]} \\
Notes: \\
- Complete table contains the following columns: \texttt{[ORG\_COLS]} \\
- Give me the answer in format \texttt{"Final Answer: True / False"} form (should be either True or False, without any explanation)
\end{tcolorbox}

\subsection{A.3. Prompt for SQL Refinement}
The prompt used for the \vmodel\ LLM to refine the SQL queries is as follows. 
\begin{tcolorbox}[colback=gray!10, colframe=black!50, title=Prompt: SQL Refinement, fonttitle=\bfseries]
\ttfamily
\small
Please refine the given SQL according to the question and the table header. \\
The SQL is: \texttt{[SQL]} \\
The table header: \texttt{[Header]} \\
The question is: \texttt{[Question]} \\
Notes: \\
- The current SQL query can not solve the problem well. \\
- Please optimize it and return the optimized SQL directly.
\end{tcolorbox}

\subsection{A.4. Prompt for Sub-table Refinement}
The prompt used for the \vmodel\ LLM to verify, and if necessary, to refine the sub-table is as follows.
 \begin{tcolorbox}[colback=gray!10, colframe=black!50, title=Prompt: \vmodel\ Post-Decomposition, fonttitle=\bfseries]
\ttfamily
\small
Please answer the question using the given table. \\
The question is: \texttt{[Question]} \\
Table Columns: \texttt{[Columns]} \\
Table Content: \texttt{[TAB\_CONTENT]} \\
CASE 1: If you can answer the question correctly, directly return the answer (without any explanation). \\
CASE 2: If information is missing, choose the necessary action with Python code for the DataFrame. \\
Complete DF column names: \texttt{[ORG\_COLS]} \\
Action list: \\
1. Retrieve\_rows (you need additional rows information to answer the question) \\
2. Retrieve\_columns (you need additional columns information to answer the question) \\
3. Retrieve\_cells (you need additional cells information to answer the question) \\
In this case, directly return the action names (can be one action or more) with their Python code, for example: \\
Action: Retrieve\_rows \\
Code: "Python\_code" \\
Notes: \\
- You should either return the answer or return the action with its Python code (no other cases).
\end{tcolorbox}

\begin{table*}[t]
\small
\centering 
\begin{tabular}{ll}
\toprule
\textbf{Purpose}  & \textbf{Regular expressions}  \\
\midrule
{\multirow{3}{*}{Column name filters}} &  \verb|1: select\s+(.*?)\s+from|\\
& \begin{lstlisting}[basicstyle=\ttfamily, numbers=none]
2:([\w\s\(\)]+)\s*(=|like|not like|in|not in|>|<|
\end{lstlisting} \\
& \begin{lstlisting}[basicstyle=\ttfamily, numbers=none]
<>|>=|<=|!=|between|is null|is not null)
\end{lstlisting} \\
\midrule
{\multirow{6}{*}{Condition \& Value filters}} & \verb|1: where\s+(.*)| \\
& \verb|2: \(\s*select.*?where\s+(.*?)\s*\)| \\
& \begin{lstlisting}[basicstyle=\ttfamily, numbers=none]
3:(\w+)\s*(=|!=|<>|>|<|>=|<=|like|not like|in|not in|
\end{lstlisting} \\
& \begin{lstlisting}[basicstyle=\ttfamily, numbers=none]
between|is null|is not null)\s*('?[\w\s\-%]+?'?)
\end{lstlisting}
 \\ 
 & \begin{lstlisting}[basicstyle=\ttfamily, numbers=none]
4:(?:(where|having)\s+(.*?))?(?:\s*group by\s+
\end{lstlisting} \\
& \begin{lstlisting}[basicstyle=\ttfamily, numbers=none]
([\w,\s]+))?(?:\s*limit\s+(\d+))?
\end{lstlisting} \\
\bottomrule
\end{tabular}
\caption{Regular expressions used to parse SQL queries.}
\label{tab:regex}
\end{table*}

\subsection{A.5. Prompt for Answer Generation}
The prompt used for the \amodel\ LLM to generate the final question answer is shown as follows.
\begin{tcolorbox}[colback=gray!10, colframe=black!50, title=Prompt: Answer Generator, fonttitle=\bfseries]
\ttfamily
\small
Please answer the question according to the given table. \\
The header is: \texttt{[Columns]} \\
The table content: \texttt{[TAB\_CONTENT]} \\
The question is: \texttt{[Question]} \\

Notes: \\
- Give me the answer in the format \texttt{"Final Answer: AnswerName1, AnswerName2..."} form (should be a number or entity names, as short as possible, without any explanation).
\end{tcolorbox}

\section{Appendix B. Additional Details on the \dmodel}
\subsection{B.1. Regex Used by the  \dmodel}\label{app:tab_decomp}
The regular expressions used by our \dmodel\ to extract the column names, conditions, and values are summarized in Table~\ref{tab:regex}.

\subsection{B.2. Effectiveness of the Regex-based Parser}
To assess the effectiveness of our regex-based parser, we empirically compare it with an LLM-based parser (\texttt{GPT-4o-mini} with the prompt shown below). We use 100 random SQL queries generated by our \smodel\ and manually evaluate whether each parser can extract the intended column names, conditions, and values.

\begin{tcolorbox}[colback=gray!10, colframe=black!50, title=Prompt: LLM-based SQL Parsing, fonttitle=\bfseries]
\ttfamily
\small
Please directly extract the column names and filter conditions in the given SQL based on the question for table decomposition. \\
Question: \texttt{[Q]} \\
SQL: \texttt{[SQL]} \\

Note: \\
- Directly return the answer, without explanation.
\end{tcolorbox}

As shown in Table~\ref{tab:parser_accuracy}, our regex-based parser achieves an accuracy of 87\%, outperforming the LLM-based parser (i.e., 70\%) substantially. The LLM-based parser often hallucinates non-existent columns and tends to omit critical components of the SQL queries. In contrast, the regex-based parser performs reliably in most cases and yields consistent results, even though it may still fail on highly complex queries not covered by the regular expressions. 

\begin{table}[ht]
\centering
\caption{Accuracy comparison between regex-based and LLM-based SQL parsers over 100 SQL queries.}
\label{tab:parser_accuracy}
\small
\begin{tabular}{lcc}
\toprule
\textbf{Parser} & \textbf{Accuracy (\%)} \\
\midrule
LLM-based Parser     & 70.0    \\
\textbf{Regex-based Parser (ours))}   & \textbf{87.0}    \\
\bottomrule
\end{tabular}
\end{table}



\begin{table*}[t]
\centering 
\small
\begin{tabular}{llll}
\toprule
& \textbf{SQL}  & \textbf{DataFrame} & \textbf{Examples}  \\
\toprule
{\multirow{6}{*}{Comparison}} &  \verb|=| & \verb|==| & \verb|df[df['age'] == 30]| \\
& \verb|!= or <>| & \verb|!=| & \verb|df[df['age'] != 30]| \\
& \verb|>| & \verb|>| & \verb|df[df['salary'] > 5000]| \\
& \verb|<| & \verb|<| & \verb|df[df['salary'] < 5000]| \\
& \verb|>=| & \verb|>=| & \verb|df[df['rating'] >= 4.5]| \\
& \verb|<=| & \verb|<=| & \verb|df[df['rating'] <= 4.5]| \\
\toprule
{\multirow{3}{*}{Logic}} &  \verb|AND| & \verb|&| & \verb|df[(df['age'] > 25) & (df['salary'] > 500]| \\
& \verb|OR| & | & \verb|df[(df['age'] > 25)| | \verb|(df['salary'] > 500)]| \\
& \verb|NOT| & \verb|~| & \verb|df[~(df['active'] == 1)]| \\
\toprule
{\multirow{6}{*}{Range}} &  \verb|BETWEEN AND| & \verb|>= x &| & \verb|df[(df['age'] >= 20) &]| \\
&  & \verb|<= y| & \verb|(df['age'] <= 30)]|\\
& \verb|IN| &  \verb|isin| & \verb|df[df['city'].isin| \\
& & & \verb|(['New York', 'Los Angeles'])]| \\
& \verb|NOT IN| & \verb|~.isin| & \verb|df[~df['city'].isin| \\
& & & \verb|(['New York', 'Los Angeles'])]| \\
\toprule
{\multirow{2}{*}{NULL}} &  \verb|IS NULL| & \verb|isna| & \verb|df[df['phone'].isna()]| \\
& \verb|IS NOT NULL| & \verb|notna| & \verb|df[df['phone'].notna()]|  \\
\toprule
{\multirow{4}{*}{Matching}} &  \verb|LIKE| & \verb|startswith| & \verb|df[df['name'].str.startswith('A')]| \\
&  & \verb|contains| & \verb|df[df['name'].str.contains('A')]|  \\
&  & \verb|endswith| & \verb|df[df['email'].str.endswith('A')]| \\
& \verb|NOT LIKE| & \verb|~.contains| & \verb|df[~df['name'].str.contains('A')]| \\
\toprule
Sorting &  \verb|ORDER BY| & \verb|sort_values| & \verb|df.sort_values('age', ascending=True/False)| \\
\toprule
{\multirow{2}{*}{Grouping}} &  \verb|GROUP BY| & \verb|groupby| & \verb|df.groupby('column')| \\
& \verb|HAVING| & \verb|filter| & \verb|df.groupby().filter(lambda x: len(x) > x)|\\
\toprule
{\multirow{5}{*}{Aggregation}} &  \verb|SUM| & \verb|sum| & \verb|df['salary'].sum()| \\
& \verb|AVG| & \verb|mean| & \verb|df['salary'].mean()|\\
& \verb|COUNT| & \verb|count| & \verb|df['salary'].count()|\\
& \verb|MIN| & \verb|min| & \verb|df['salary'].min()|\\
& \verb|MAX| & \verb|max| & \verb|df['salary'].max()|\\
\toprule
{\multirow{2}{*}{LIMIT}} &  \verb|LIMIT| & \verb|head| & \verb|df.head(10)| \\
& \verb|OFFSET LIMIT| & \verb|iloc| & \verb|df.iloc[5:15]|\\
\toprule
\end{tabular}
\caption{Mapping between SQL operators and Python
DataFrame operations.}
\label{tab:sql_df}
\end{table*}

\begin{figure*}[t]
    \centering
    \begin{subfigure}[c]{0.41\textwidth}
        \centering
        \includegraphics[width=1\linewidth]{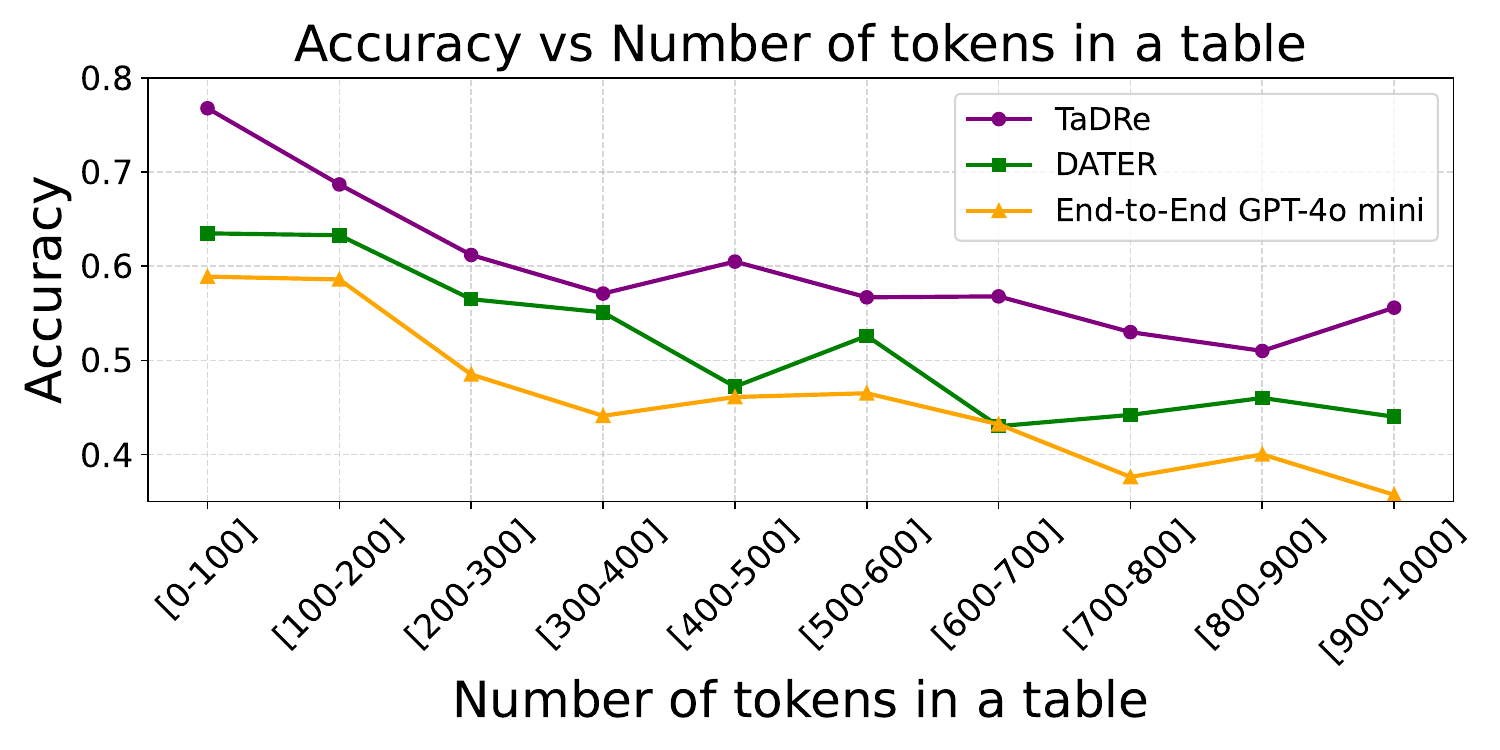}
        \caption{\texttt{\wdataset}}
        \label{fig:token_model_acc}
    \end{subfigure}
    \begin{subfigure}[c]{0.41\textwidth}
        \centering
        \includegraphics[width=1\linewidth]{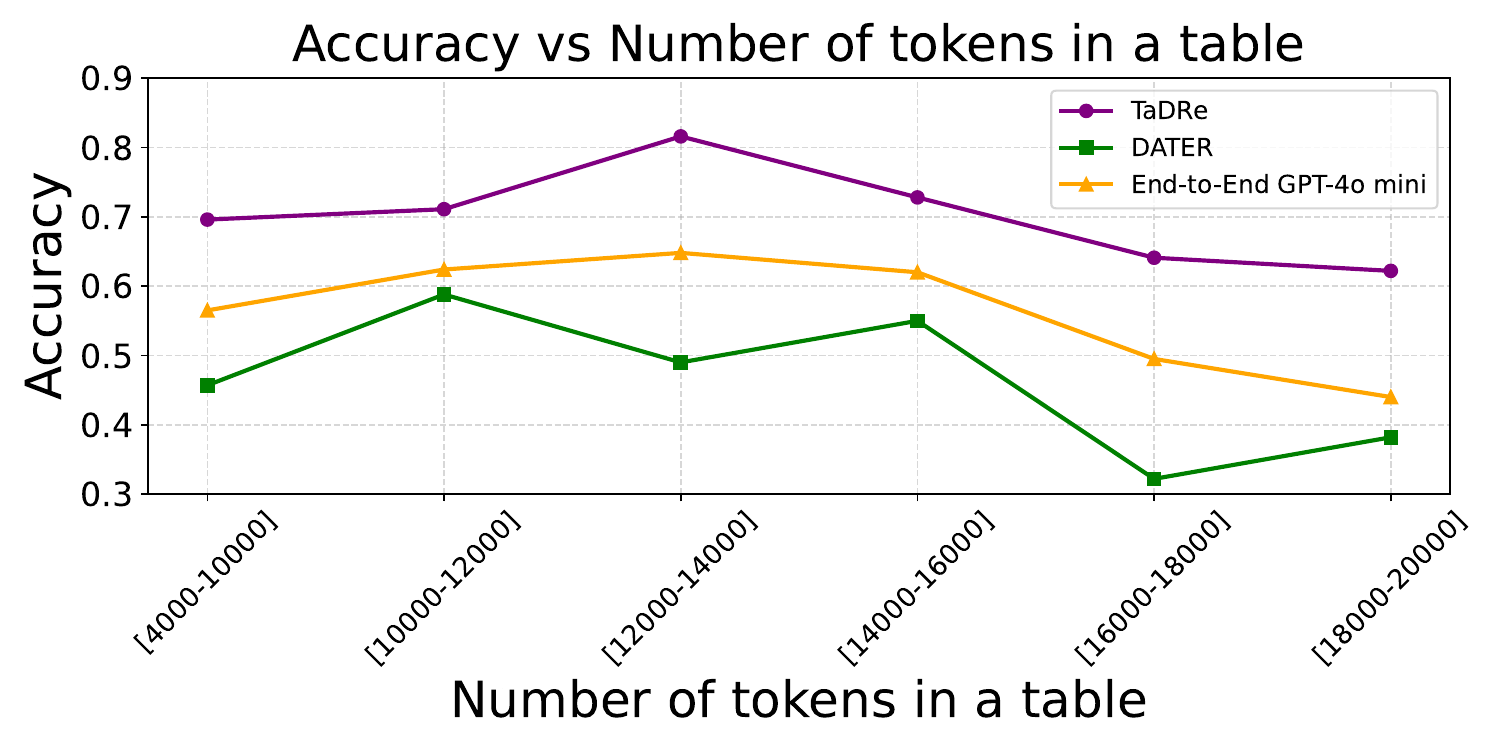}
        \caption{\texttt{\sdataset}}
        \label{fig:token_large_acc}
    \end{subfigure}
   \caption{Breakdown of model accuracy over tables of different sizes (using \texttt{GPT-4o mini} as the backbone). }
    \label{fig:token_vs_models}
\end{figure*}

\subsection{B.3. SQL-to-Python Mapping}
The \dmodel\ maps SQL operators to Python DataFrame operations following Table~\ref{tab:sql_df}.

\subsection{B.4. \model\ Performance by Table Size}
We show the accuracy of \model\ and two best baselines, DATER and End-to-End \texttt{GPT-4o mini}, over tables of different sizes in Figure~\ref{fig:token_vs_models}. The accuracy drops as the tables get larger, which is expected. Importantly, \model\ drops the slowest, demonstrating its stronger robustness and scalability towards large tables. 

\begin{figure}[t]
     \centering
     \includegraphics[width = 1\linewidth]{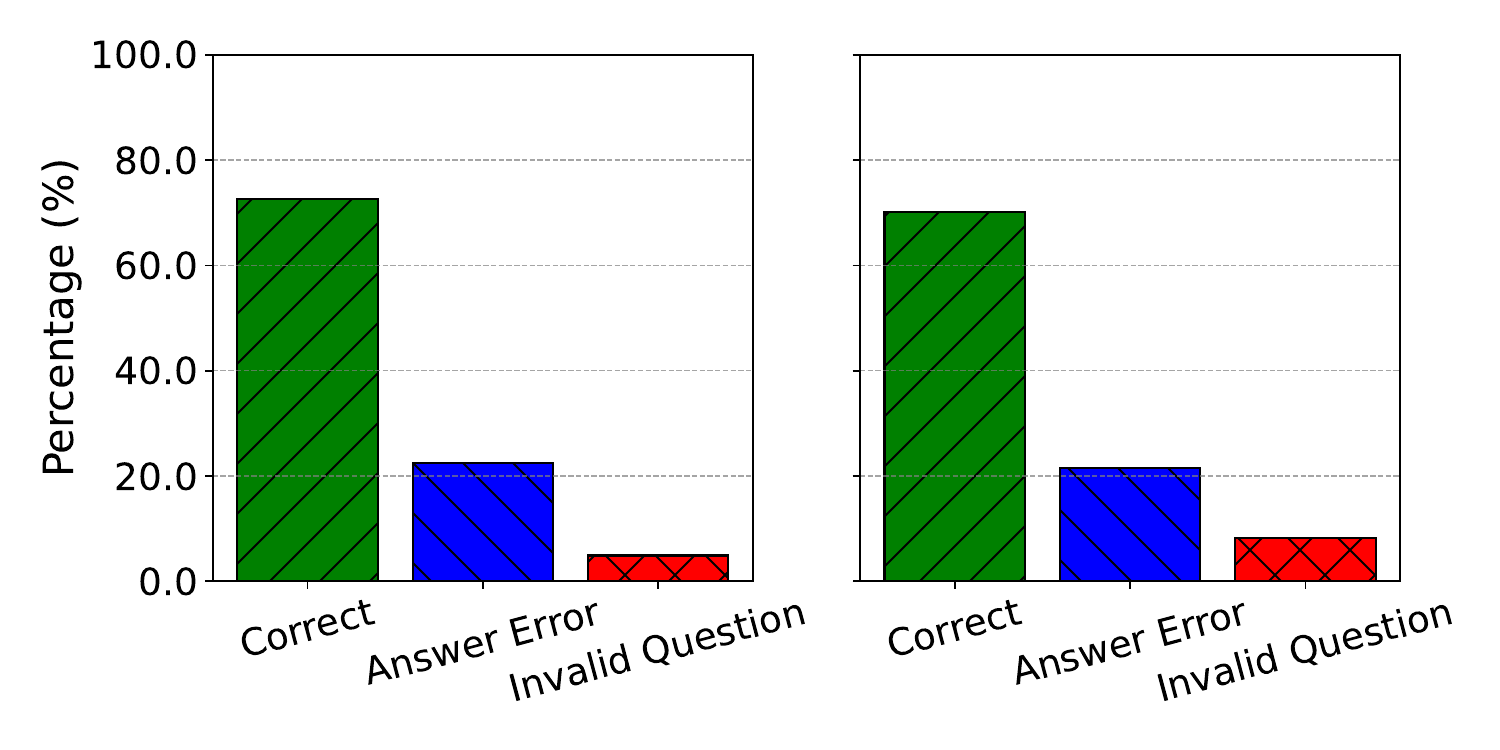}
     \caption{Error analysis for our LLM-based QA pair generation process: (a) Results for the \texttt{\sdataset} dataset; (b) Results for the \texttt{\ldataset} dataset. ``Correct'' means that the QA pairs are correct. ``Answer Error'' means that the generated questions are valid, but the answers are incorrect. ``Invalid Question'' means that the questions themselves are invalid.}
     \label{fig:qa_analysis}
\end{figure}

\section{Appendix C. Additional Details on Dataset Construction}
\subsection{C.1. QA Pair Generation Errors}\label{app:dataset}
Figure~\ref{fig:qa_analysis} summarizes error statistics of our LLM-based QA pair generation process. Particularly, for the ``Answer Error'' cases, the generated questions are valid while the answers are incorrect. We manually correct the answers according to the tables to retain more QA pairs in the final datasets. For the ``Invalid Question'' cases, we remove the questions and their corresponding answers. 

\subsection{C.2. Prompt for QA Pair Generation}
The prompts used to generate QA pairs are as follows.
\begin{tcolorbox}[colback=gray!10, colframe=black!50, title=(a) Cell-based, fonttitle=\bfseries]
\ttfamily
\small
Use the given table as evidence. \\
Table Header: \texttt{[HEADER]} \\
Table Content: \texttt{[TAB]} \\
Task: Randomly select one cell in the table as the answer and generate a question that can produce the answer. \\
Return both the question and the answer. Repeat this process 10 times and return to me 10 QA pairs. \\
Notes: \\
- Directly return questions in format \texttt{"Q: question\_content; A: answer\_content"} (without any explanation). \\
- Try to make the question diverse. \\
- Keep answers as concise as possible and only contain entities.
\end{tcolorbox}

\begin{tcolorbox}[colback=gray!10, colframe=black!50, title=(b) Row-based, fonttitle=\bfseries]
\ttfamily
\small
Use the given table as evidence. \\
Table Header: \texttt{[HEADER]} \\
Table Content: \texttt{[TAB]} \\
Selected Row: \texttt{[ROW]} \\
Task: Please use the selected row above to generate one question using the information within this row. Return the question and its answer. Repeat this process 4 times. \\
Notes: \\
- Directly return questions in format \texttt{"Q: question\_content; A: answer\_content"} (without any explanation). \\
- Try to make the question diverse. \\
- Do not include "selected row/given data" in questions, as the selected columns are not known to the person answering the question. \\
- Keep answers as concise as possible and only contain entities.
\end{tcolorbox}

\begin{tcolorbox}[colback=gray!10, colframe=black!50, title=(c) Column-based, fonttitle=\bfseries]
\ttfamily
\small
Use the given table as evidence. \\
Table Header: \texttt{[HEADER]} \\
Table Content: \texttt{[TAB]} \\
Selected Column: \texttt{[COL]} \\
Task: Please use the selected column above to generate one question using the information within this column. Return the question and its answer. Repeat this process 4 times. \\
Notes: \\
- Directly return questions in format \texttt{"Q: question\_content; A: answer\_content"} (without any explanation). \\
- Try to make the question diverse. \\
- Do not include "selected column/given data" in questions, as the selected columns are not known to the person answering the question. \\
- Keep answers as concise as possible and only contain entities.
\end{tcolorbox}

\begin{tcolorbox}[colback=gray!10, colframe=black!50, title=(d) Sub-table-based, fonttitle=\bfseries]
\ttfamily
\small
Use the given table as evidence. \\
Table Header: \texttt{[HEADER]} \\
Table Content: \texttt{[TAB]} \\
Sub-table: \texttt{[SUB]} \\
Task: Please use the sub-table above to generate one question using the information within this sub-table. Return the question and its answer. Repeat this process 4 times. \\
Notes: \\
- Directly return questions in format \texttt{"Q: question\_content; A: answer\_content"} (without any explanation). \\
- Try to make the question diverse. \\
- Do not include "sub-table/given data" in questions, as the sub-table is not known to the person answering the question. \\
- Keep answers as concise as possible and only contain entities.
\end{tcolorbox}

\subsection{C.3. Prompt for Table Expansion}
The prompt used to expand a small table into a larger one is shown as follows.
\begin{tcolorbox}[colback=gray!10, colframe=black!50, title=Prompt: Table Expansion via Synthesis, fonttitle=\bfseries]
\ttfamily
\small
Please use the synthesized data to expand the existing table and increase its rows and columns. \\
Table Header: \texttt{[HEADER]} \\
Table Content: \texttt{[TAB]} \\

Notes: \\
- Directly returns the new table. \\
- Make sure there are no duplicate rows and columns.
\end{tcolorbox}

\subsection{C.4. Executable SQL Queries}
\label{sec:sql_exec_def}

Let $T$ be a relation instance (i.e., a table) with known column names and data values. Let $\text{Cols}(T)$ be the set of column names in $T$. A SQL query $s$ is considered \textit{executable} on $T$ if the following conditions are met: 

\begin{itemize}
    \item \textbf{1. Column Validity:} All column references in $s$ (including those used in projection $\pi$, selection $\sigma$, renaming $\rho$, or extended relational operations) correspond to existing column names in $T$. 
    
    \item \textbf{2. Semantic Interpretability:} Every operation in $s$ can be semantically interpretable and safely applied to the table's values. This includes:
    \begin{itemize}
        \item Value-level operations (e.g., arithmetic, comparison, and string functions).
        \item Aggregation and grouping operations (e.g., \texttt{SUM} and \texttt{GROUP BY}).
        \item Logical and set-level operations (e.g., \texttt{AND}, \texttt{IN}, \texttt{EXISTS}, and \texttt{UNION}).
    \end{itemize}

    \item \textbf{3. Algebraic Mapping:} Query $s$ can be faithfully translated into a relational algebra expression $A(s)$ whose evaluation on the instance $R$ represented by table $T$ is well-defined and returns a valid relation.
    
    \item \textbf{4. Runtime Error-Free:} The execution of $s$ on $T$ does not result in any runtime errors, such as division by zero, null dereferencing, invalid function domains, or other data-dependent failures that cannot be detected statically.
\end{itemize}

\subsection{C.5. SQL Executable Questions}
Let $q$ be a natural language question. Let $\mathsf{EXEC}^{T}_{\text{sql}}$ denote the set of all executable SQL queries on table $T$ under the above semantics. Let $M$ denotes a mapping mechanism from a question $q$ to sets of semantically equivalent SQL queries.

We say that $q$ is \textit{SQL-executable} under semantics $T$ if and only if:
\[
M(q) \cap \mathsf{EXEC}^{T}_{\text{sql}} \neq \emptyset
\]

That is, at least one semantically correct SQL query mapped from $q$ is executable on table $T$.

\subsection{C.6. Prompt for Checking Question Executability}
The prompt used for checking if a question is SQL-executable is shown as follows.

\begin{tcolorbox}[colback=gray!10, colframe=black!50, title=Prompt: SQL-executable Question Checking, fonttitle=\bfseries,
breakable]
\ttfamily
\small
Task: Given a natural language question \texttt{q} and a table column name \texttt{C}, determine whether \texttt{q} is a SQL-executable question under the following semantics: \\

A question \texttt{q} is considered SQL-executable if there exists at least one SQL query \texttt{s} that: \\
- Maps from \texttt{q} through a valid equal semantic mapping mechanism (i.e., $\texttt{s} \in \texttt{M(q)}$, and \\
- Satisfies all the following conditions: \\

1. \textbf{Column Validity}: All column references in \texttt{s} refer to column names in \texttt{C}. \\
2. \textbf{Semantic Interpretability}: Every operation in \texttt{s} — including value-level operations (e.g., comparisons, arithmetic, string functions), aggregate and grouping operations (e.g., SUM, GROUP BY), and logical/set-level operations (e.g., AND, IN, EXISTS, UNION) — can be semantically interpreted and safely applied to the actual values in the referenced columns. \\
3. \textbf{Algebraic Mapping}: \texttt{s} can be faithfully translated into a well-defined relational algebra expression whose evaluation on the table is valid and returns a relation. \\
4. \textbf{Runtime Error-Free}: The execution of \texttt{s} on the given table would not lead to runtime errors (e.g., division by zero, null dereferencing, invalid function usage, or other data-dependent failures). \\

Few-shot Examples: \\

\#\#\#\# Condition 1: Violates Column Validity \\

\textbf{Question}: \\
What is the average employee rating by department? \\

\textbf{Table Header}: \\
\texttt{["Department", "Employee Name", "Performance Score", "Join Date"]} \\

\textbf{Sample Row Values}: \\
\texttt{[["HR", "Alice", 4.5, "2018-01-01"], ["Engineering", "Bob", 3.8, "2020-06-01"]]} \\

\textbf{Explanation}: \\
Although the question mentions "department" and "employee rating", the column "employee rating" does not exist. There is no column with a matching name or sufficiently clear mapping. "Performance Score" might be intended, but the mapping is ambiguous without schema-level aliases. \\

\textbf{Condition Violated}: Column Validity \\
\textbf{Final Answer}: No \\

\#\#\#\# Condition 2: Violates Semantic Interpretability \\

\textbf{Question}: \\
What is the median value among departments? \\

\textbf{Table Header}: \\
\texttt{["Employee ID", "Name", "Department", "Salary"]} \\

\textbf{Sample Row Values}: \\
\texttt{[["E001", "Alice", "HR", 70000], ["E002", "Bob", "Engineering", 75000]]} \\

\textbf{Explanation}: \\
"Median" is a statistical operation valid only on numeric or ordered data. "Department" is a categorical string column, so computing a median on it is semantically meaningless, even if syntactically allowed. \\

\textbf{Condition Violated}: Semantic Interpretability \\
\textbf{Final Answer}: No \\

\#\#\#\# Condition 3: Violates Algebraic Mapping \\

\textbf{Question}: \\
Which students performed better than average in subjects they are strongest at? \\

\textbf{Table Header}: \\
\texttt{["Student", "Math Score", "English Score", "Science Score"]} \\

\textbf{Sample Row Values}: \\
\texttt{[["Alice", 88, 76, 91], ["Bob", 90, 85, 80], ["Charlie", 72, 95, 70]]} \\

\textbf{Explanation}: \\
The question requires identifying the strongest subject (highest score) per student, then comparing it against that subject's average across all students. This involves row-wise dynamic column selection and per-row conditional aggregation, which cannot be represented in standard relational algebra or SQL without structural transformations like unpivoting. Therefore, a valid relational algebra mapping is not possible. \\

\textbf{Condition Violated}: Algebraic Mapping \\
\textbf{Final Answer}: No \\

\#\#\#\# Condition 4: Violates Runtime Error-Free \\

\textbf{Question}: \\
What is the highest average revenue per unit cost? \\

\textbf{Table Header}: \\
\texttt{["Product", "Revenue", "Unit Cost"]} \\

\textbf{Sample Row Values}: \\
\texttt{[["P1", 10000, 0], ["P2", 20000, 100]]} \\

\textbf{Explanation}: \\
While the SQL query \texttt{SELECT MAX(Revenue / Unit Cost)} is syntactically and semantically valid, it can result in a division by zero runtime error when Unit Cost is 0, which is highly probable in the given data. Thus, it fails the Runtime Error-Free condition. \\

\textbf{Condition Violated}: Runtime Error-Free \\
\textbf{Final Answer}: No \\

Input: \\
Question: \texttt{[Q]} \\
Table Header (Column Names Only): \texttt{[C]} \\
Sample Row Values (Not the entire table): \texttt{[V]} \\

Output: \\
Is the question SQL-executable with respect to the given table? (Yes or No) \\
If Yes, explain briefly what a possible executable SQL query might look like. \\
If No, specify which condition(s) in the definition are violated. \\

Notes: \\
- For condition: Runtime Error-Free, don't assume that this condition has been violated unless the probability of \texttt{s} execution error is very high. \\
- If No, firstly specify which condition(s) in the definition are violated, and then give answers. If Yes, directly give answers. \\
- Then respond in format: \texttt{Final Answer: Yes/No}
\end{tcolorbox}

\begin{figure}[t]
     \centering
     \includegraphics[width=0.8\linewidth]{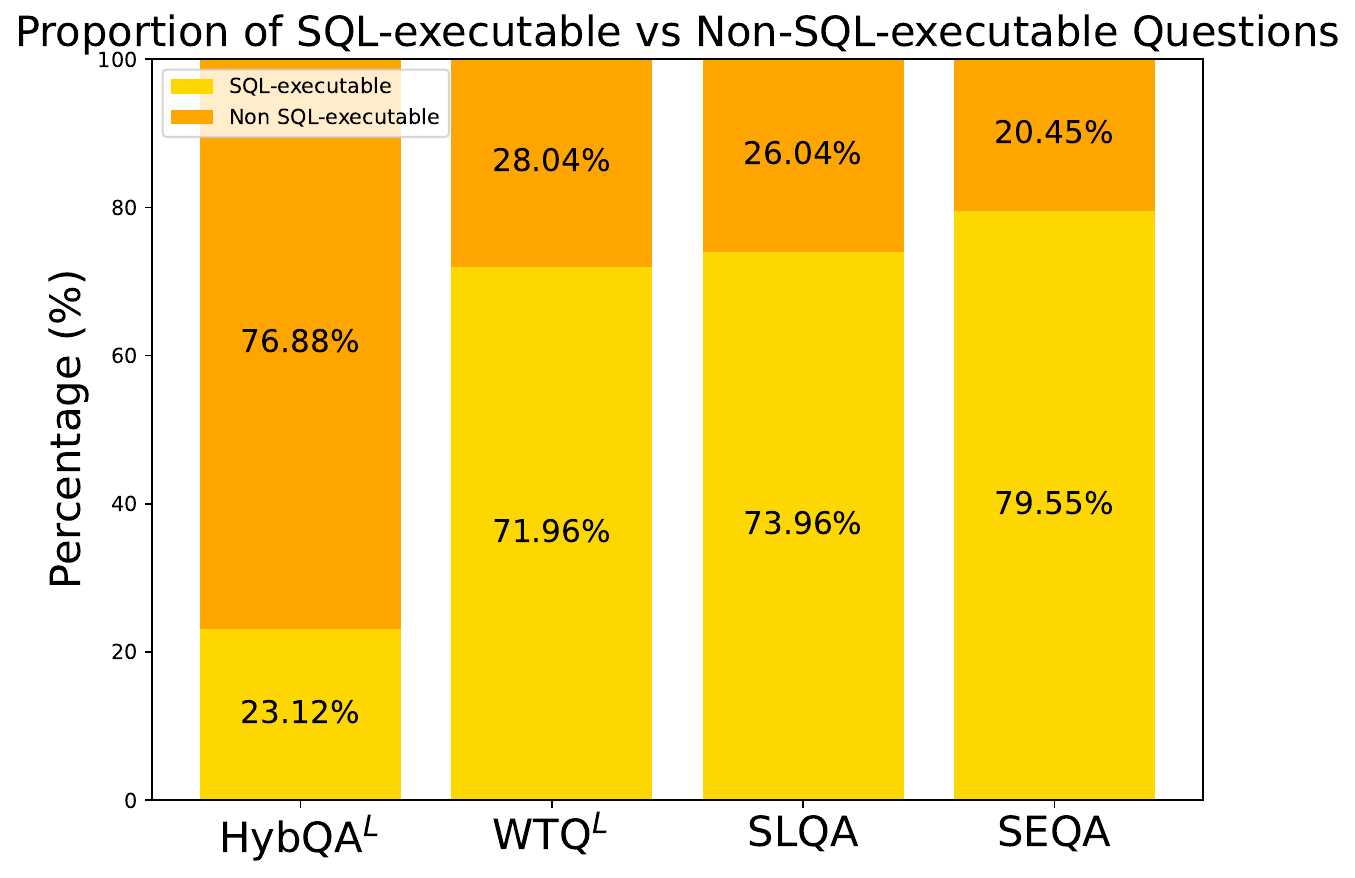}
     \caption{Question type statistics.}
     \label{fig:q_stats}
\end{figure}

Figure~\ref{fig:q_stats} shows the proportion of SQL-executable and non-SQL-executable questions in the large-table datasets. Our datasets \texttt{\sdataset} and \texttt{\ldataset} closely resemble the human-annotated dataset \texttt{\wdataset} in the question type distributions. 

We also manually check the classification accuracy of the prompt above. As Table~\ref{tab:ds_cls_acc} shows, the LLM can clarify the questions at 79\%  accuracy.

\begin{table}[t]
\centering
\small
\begin{tabular}{lc}
\toprule
\textbf{Executable SQL Condition} & \textbf{Accuracy} \\
\midrule
Column validity & 83\% \\
Semantic interpretability & 91\% \\
Algebraic mapping & 95\% \\
Runtime error-free & 90\% \\\hline
Overall & 79\% \\
\bottomrule
\end{tabular}
\caption{Question type classification accuracy.}
\label{tab:ds_cls_acc}
\end{table}

\section{Appendix D. Additional Experimental Results}
\subsection{D.1. Prompt for LLM-based Table Decomposer}\label{app:exps}
In Table~\ref{tab:enhancement} earlier,  we compared the \dmodel\ with an LLM-based table decomposition method. Here, we show the prompt used for the method as follows. 
\begin{tcolorbox}[colback=gray!10, colframe=black!50, title=Prompt: LLM Table Decomposer, fonttitle=\bfseries]
\ttfamily
\small
Please return the row and column indices of the given table that contains the information to answer the question. \\
Table header: \texttt{[Header]} \\
Table content: \texttt{[Table]} \\
Question: \texttt{[Q]} \\

Notes: \\
- Include all the information regarding the question, not just the answer. \\
- Directly return the row indices in one list and return the column indices in another list (without any explanation).
\end{tcolorbox}

\begin{figure}[t]
     \centering
     \includegraphics[width = 1\linewidth]{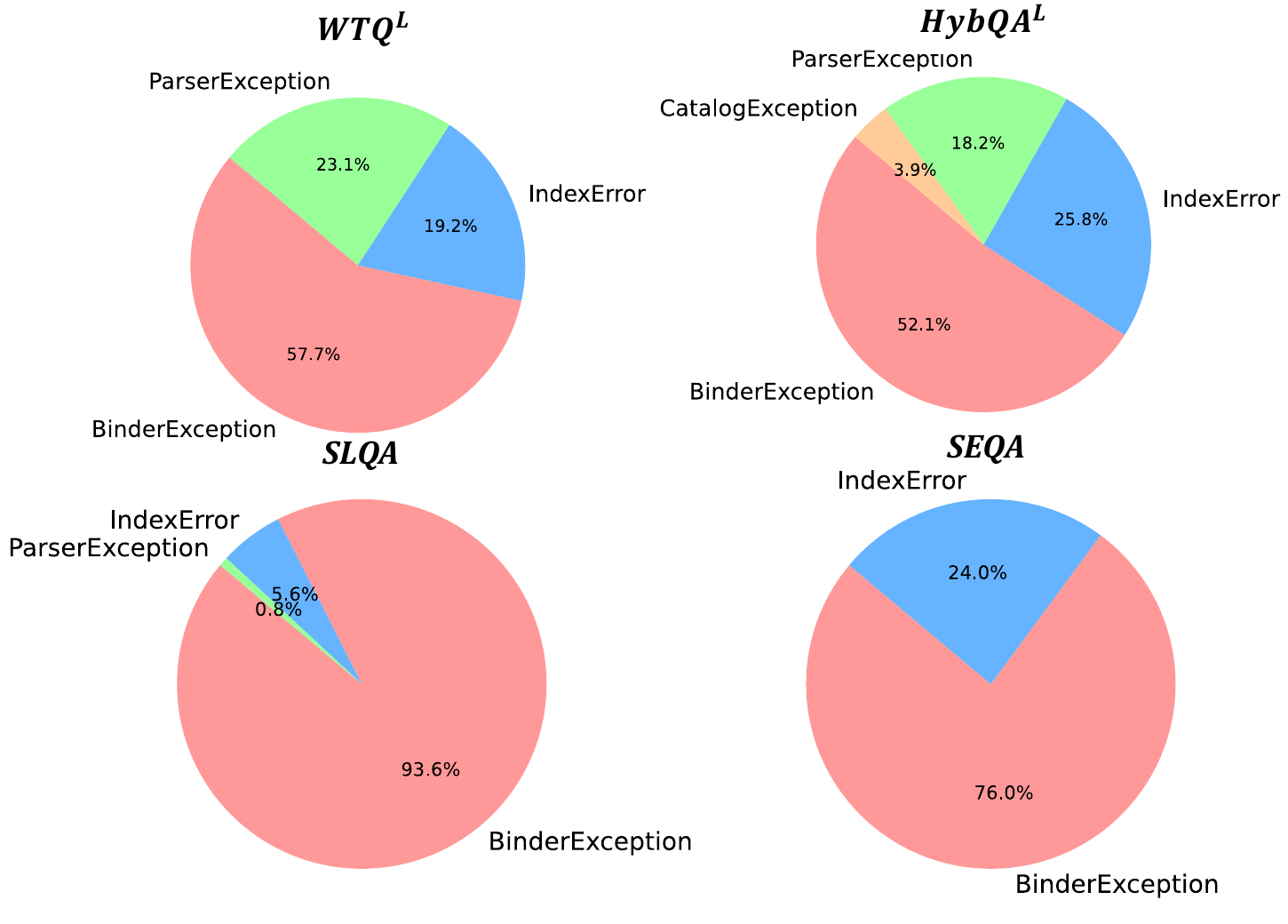}
     \caption{Statistics on SQL execution errors.}
     \label{fig:sql_errors}
\end{figure}

\subsection{D.2. SQL Execution Error Analysis}\label{app:error_analysis}
Figure~\ref{fig:sql_errors} presents statistics on errors when directly executing the \smodel's generated SQL queries to obtain answers. \texttt{BinderException}, which occurs when a query refers to invalid tables, columns, aliases, or incorrect conditions, is the most common error type. It is followed by \texttt{IndexError}, which happens at runtime due to out-of-bounds index access. The dominance of \texttt{BinderException} suggests schema or condition errors, while \texttt{IndexError} indicates issues with indexing. 

On different datasets, the distributions of errors vary. This is because different datasets have various table size distributions (see Table~\ref{tab:dataset-features}), reflected in the number of rows and columns, and the complexity of the contents in each cell, e.g., the presence of lengthy descriptive texts in \texttt{\hdataset}$^L$.

\begin{table}[t]
\centering
\begin{tabular}{lc}
\toprule
\textbf{Fail Case} & \textbf{Percentage} \\ 
\midrule
{Decompose Fail} & 28.33\% \\
{Python Retrieval Fail} & 31.67\% \\ 
{Decompose + Retrieval Fail} & 16.67\% \\
{Reasoning Fail with Valid Sub-tables}  & 83.33\% \\
\bottomrule
\end{tabular}
\caption{Failure rates for different error types.}
\label{tab:fail_types}
\end{table}

\subsection{D.3. Error Analysis for Modules of \model}
In addition to SQL execution errors, we analyze errors in other stages of the \model. There are three main types of errors, as summarized in Table~\ref{tab:fail_types} over 60 error cases. 
``Decompose Fail'' means that the decomposition process either fails to extract any useful information (falls back to using the entire table) or the resulting sub-table lacks the information necessary to answer the question. ``Python Retrieval Fail'' (for the CoTR process)  means that either Python execution failure or the Python program fails to retrieve useful information. ``Reasoning Fail with Valid Sub-tables'' indicates that the final output errors are caused by the LLM reasoning process, even when the given sub-tables are valid. 
The main reason for errors, as shown in the table, is the limited LLM reasoning capability, which caused  $83.33\%$ of all errors, while only $16.67\%$ of the errors are due to the decomposition and retrieval failure of \model --i.e., the final sub-table lacks the necessary information to derive the correct answer.

\subsection{D.4. Case Study}\label{app:case_study}
\begin{figure*}[ht!]
    \centering
    \begin{subfigure}[b]{0.48\linewidth}
        \centering
        \includegraphics[width=\linewidth]{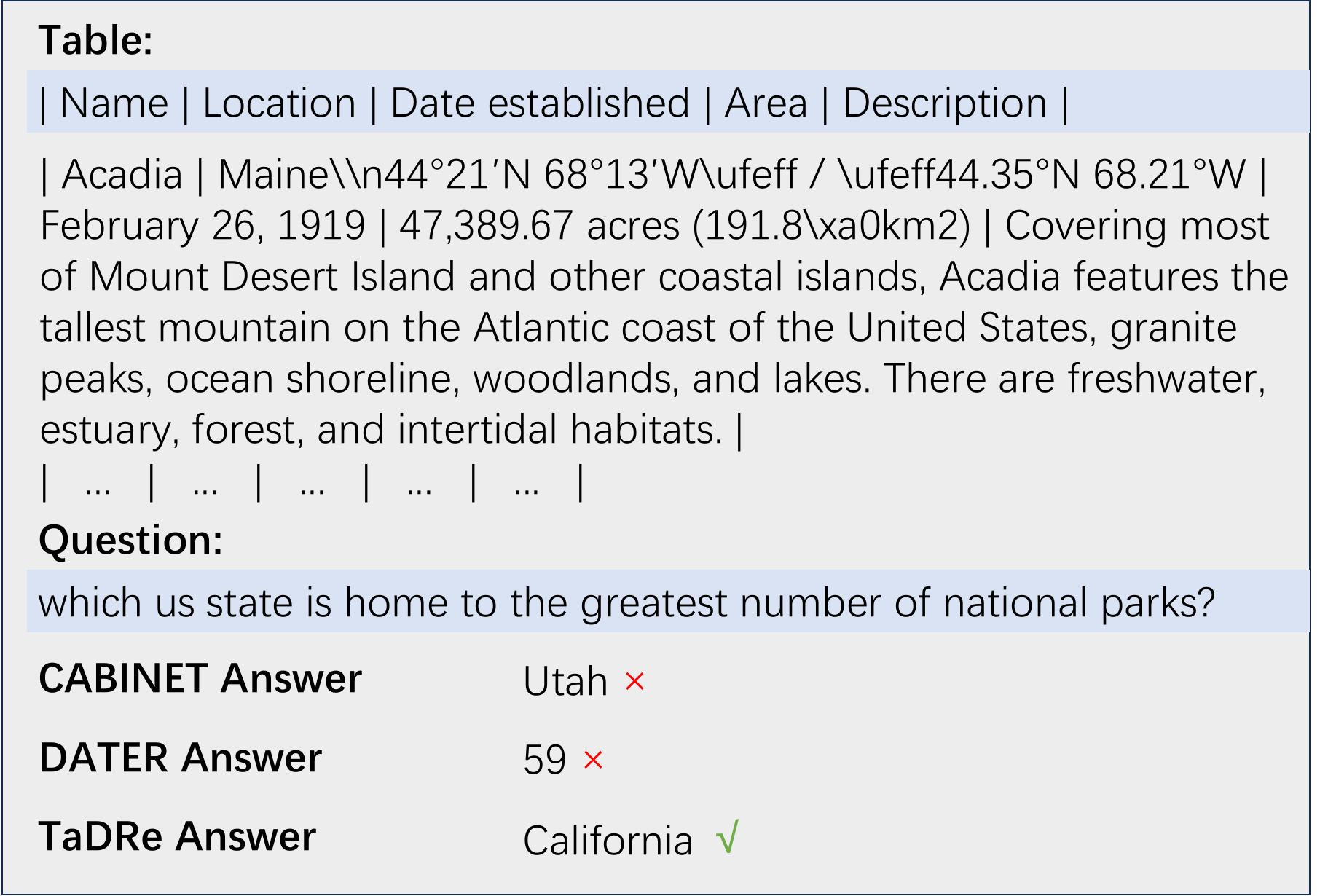}
       \caption{}
        \label{fig:case1}
    \end{subfigure}
    \begin{subfigure}[b]{0.48\linewidth}
        \centering
        \includegraphics[width=\linewidth]{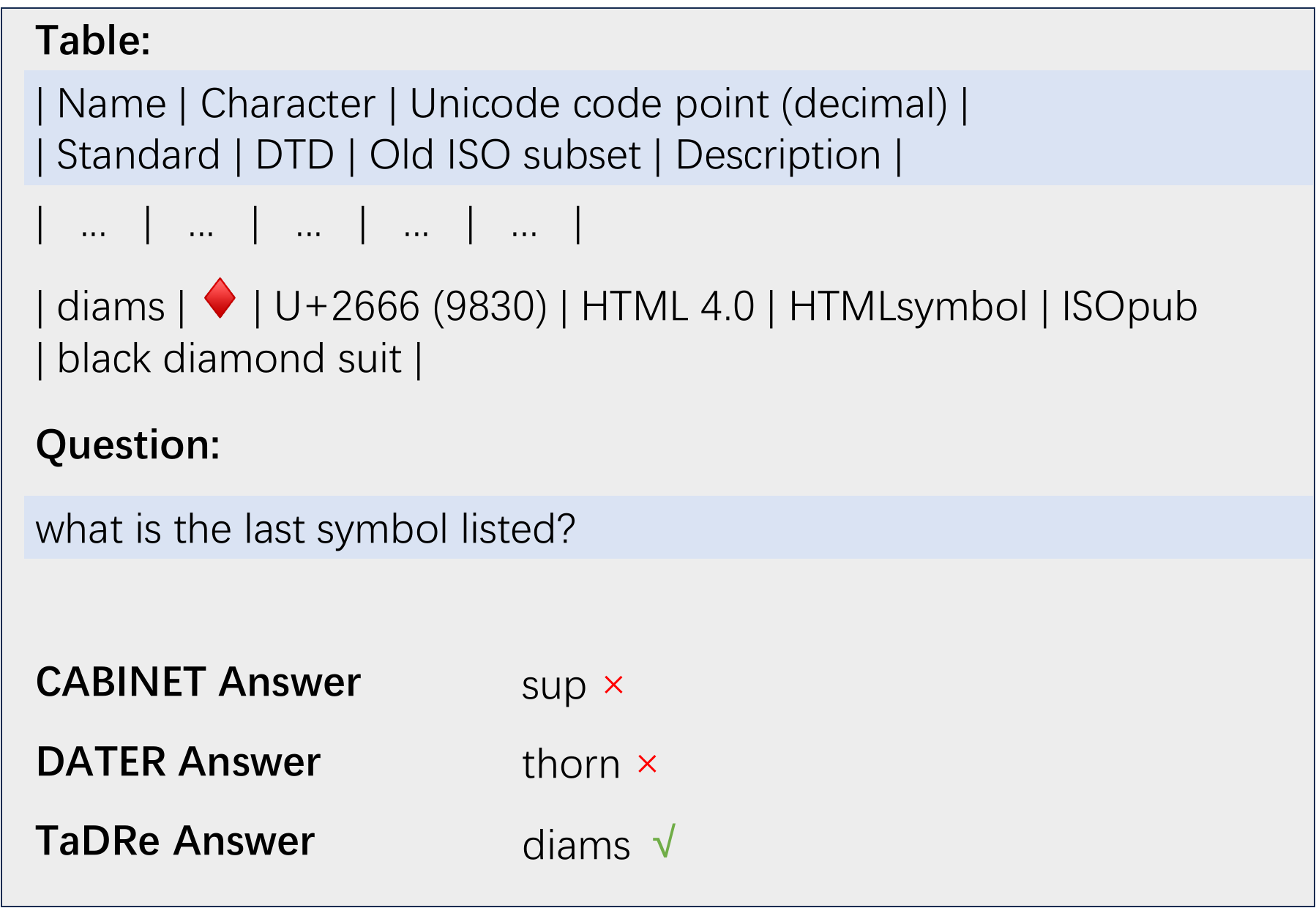}
      \caption{}
        \label{fig:case2}
    \end{subfigure}
    \caption{Two TableQA questions that \model\ answers correctly while CABINET and DATER fail.}
    \label{fig:case_study}
\end{figure*}

\begin{figure*}[t]
     \centering
     \includegraphics[width=0.9\linewidth]{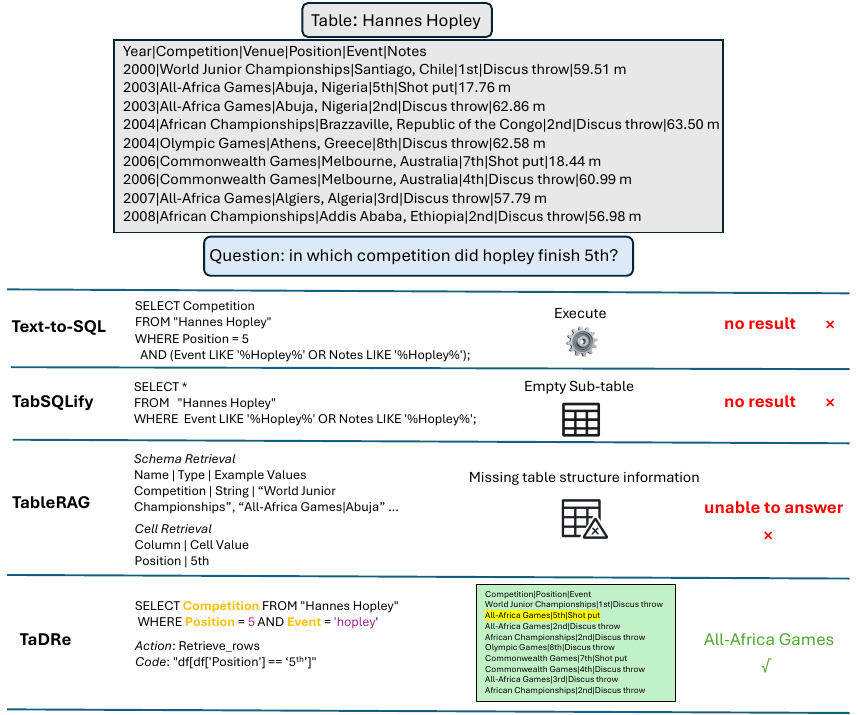}
     \caption{A question that \model\ answers correctly while program-as-decomposer models fail.}
     \label{fig:err_case_llm}
\end{figure*}

We show three examples in Figure~\ref{fig:case_study} and Figure~\ref{fig:err_case_llm}, where \model\ successfully returns the question answers, while existing table decomposition-based models fail. CABINET fails to accurately identify and assign higher weights to the cells containing the answers when handling large tables, while DATER may miss the answer field during the table decomposition process. Reasons why the program-as-decomposer baselines fail are shown in Figure~\ref{fig:case_study}.

\end{document}